\documentclass[11pt]{article}

\usepackage{microtype} 
\usepackage{booktabs}  
\usepackage{url}  
\usepackage{todonotes}
\usepackage{amsmath}
\usepackage{amsthm}
\usepackage{subcaption}
\usepackage{lipsum}
\usepackage{bbm}
\usepackage{listings}
\usepackage{xcolor}
\usepackage{threeparttable}

\newcommand{\F}{\mathcal{F}}
\newcommand{\RI}[1]{\mathit{RI}_{#1}}

\newcommand{\xhack}{\mathrm{XHack}}
\newcommand{\phifk}{\phi^{f,k}}

\definecolor{keywordblue}{RGB}{0,0,180}
\definecolor{commentgreen}{RGB}{0,140,0}
\definecolor{stringred}{RGB}{180,0,0}

\usepackage[preprint]{automl}
\usepackage{natbib}
\newtheorem*{definition}{Definition}

\title{Equally Good, Yet Different: Benchmarking Rashomon sets in AutoML packages}

\author[1,2]{\nameemail{Katarzyna Woźnica}{katarzyna.woznica@pw.edu.pl}}
\author[1]{\nameemail{Katarzyna Rogalska}{katarzyna.rogalska2.stud@pw.edu.pl}}
\author[1]{\nameemail{Zuzanna Sieńko}{zuzanna.sienko.stud@pw.edu.pl}}
\author[3]{\nameemail{Mustafa Cavus}{mustafa.cavus@eskisehir.edu.tr}}

\affil[1]{Warsaw University of Technology}
\affil[2]{Systems Research Institute, Polish Academy of Sciences}
\affil[3]{Eskisehir Technical University}

\hypersetup{%
  pdfauthor={}, 
  pdftitle={},
  pdfsubject={},
  pdfkeywords={Rashomon set, predictive multiplicity, AutoML, model multiplicity, x-hacking}
}

\begin{document}

\maketitle

\begin{abstract}
The Rashomon effect describes the existence of multiple near-optimal models that achieve comparable performance while offering fundamentally different explanations. This creates a critical vulnerability in AutoML: x-hacking, the selective post-hoc choice of a model based on its explanation rather than predictive merit. No existing AutoML framework exposes this risk. We introduce ARSA ML, an open-source Python framework that quantifies Rashomon set structure and predictive multiplicity within AutoML pipelines. Using ARSA ML, we benchmark AutoGluon and H2O across 28 binary classification datasets, and conduct a post-hoc x-hacking analysis revealing a consistent structural asymmetry: AutoGluon produces larger, diverse sets with stable explanations, while H2O generates compact sets with markedly higher prediction divergence and explanation instability — making H2O users considerably more exposed to x-hacking. This gap persists across all evaluated metrics and epsilon thresholds, pointing to a fundamental difference in each framework's model-building strategy. ARSA ML is available at \url{https://pypi.org/project/arsa-ml/}.
\end{abstract}

\section{Introduction}

The Rashomon effect \citep{breiman_rashomon_effect} describes multiple predictive models that achieve comparable performance while explaining the same phenomenon in fundamentally different ways. This occurs because data provide only an imperfect representation of reality \citep{ganesh2025systemizing}—an effect amplified by variations in training sets \citep{renard2024understanding}, hyperparameters \citep{cavus2025role}, and preprocessing \citep{cavus2025investigating}. Consequently, this leads to model multiplicity at the dataset level and predictive multiplicity at the individual level~\citep{damour, marxpredictive}, where near-optimal models may issue conflicting predictions in high-stakes domains such as credit scoring or medical diagnosis. Characterizing the Rashomon set \citep{variable_importance_rashomon_set}—the collection of models within a specified performance tolerance—is therefore essential to understanding these implications.

Because near-optimal models may assign different importance to different features~\citep{rudinAmazingThingsCome2024}, the Rashomon set induces a corresponding multiplicity of explanations. For instance, a model attributing loan rejection primarily to income can coexist with one attributing it to credit history, both achieving identical predictive performance. This explanation multiplicity enables x-hacking~\citep{sharma2024x}: the selective, post-hoc choice of a model from the Rashomon set based on its explanation properties rather than predictive merit. The susceptibility of a model selection process to x-hacking—which we call \emph{x-hackability}—therefore depends on how large and how explanation-diverse its Rashomon set is.

The Rashomon perspective is largely absent from Automatic Machine Learning (AutoML) systems~\citep{AutoML,agtabular,autosklearn,h2oautoml}. By optimizing a single performance criterion and returning one best model, AutoML creates the illusion of a uniquely correct solution—particularly misleading for users with limited ML expertise who are least equipped to recognize the existence of equally valid alternatives. While AutoML does engage with model diversity in ensemble construction, this serves error decorrelation rather than acknowledging that multiple models explain the phenomenon differently. \citet{sharma2024x} has demonstrated x-hacking's feasibility in auto-sklearn \citep{autosklearn} pipelines, but a more fundamental question remains: do AutoML systems structurally vary in the x-hacking opportunities they provide, and do they offer users any means to recognize this risk? To our knowledge, no existing AutoML framework exposes Rashomon set metrics, quantifies explanation disparity across near-optimal models, or alerts users to output x-hackability.

\begin{figure}
    \centering
    \includegraphics[width=0.7\linewidth]{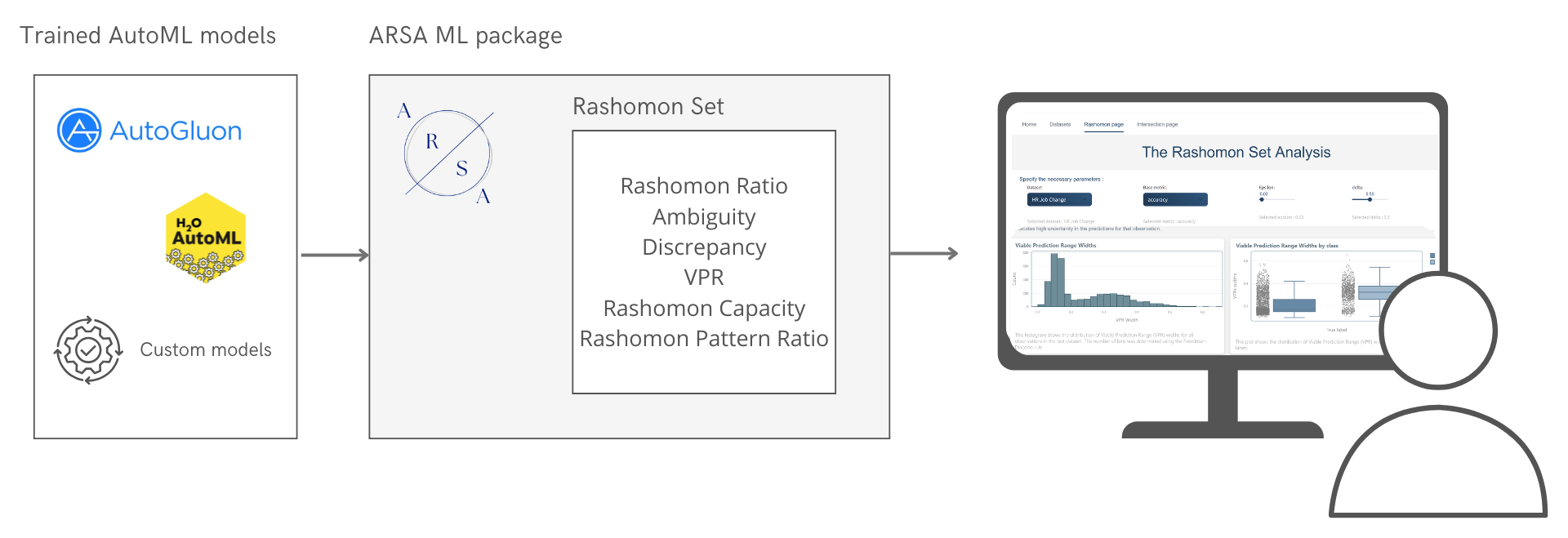}
    \caption{General schema of using the ARSA ML package. It summarizes the Rashomon sets for two packages: AutoGluon and H2O, but there is a possibility to provide results from any AutoML package.}
    \label{fig:package_overview}
\end{figure}
To address this gap, we introduce \textbf{ARSA ML} (AutoML Rashomon set Analysis), a framework for analyzing the Rashomon effect within AutoML pipelines (see Figure~\ref{fig:package_overview}). ARSA ML provides unified implementations of Rashomon set metrics~\citep{marxpredictive, semanova_noise, rashomon_ratio_volume, probab, capacity} spanning three levels: set-level metrics characterizing size and behavioral diversity, dataset-level metrics quantifying predictive multiplicity, and instance-level metrics identifying observations most exposed to conflicting predictions. ARSA ML also introduces the \emph{Rashomon Intersection}—the overlap of Rashomon sets constructed under multiple evaluation criteria—together with a principled multi-criterion reference model selection methodology. The framework offers a common interface for AutoGluon \citep{agtabular} and H2O \citep{h2oautoml}, with a converter for other frameworks.

We apply the ARSA ML package to benchmark AutoGluon and H2O across Rashomon diversity, and investigate the following research questions:  \textbf{RQ1}: How do their distinct model-building strategies affect Rashomon set size and diversity? \textbf{RQ2}: To what extent do the frameworks exhibit ambiguity, discrepancy, viable prediction range, and Rashomon capacity for identical observations? \textbf{RQ3}: Do their near-optimal models converge on the same influential features, or does explanation disparity make one framework more susceptible to x-hacking?

\section{Related Work}

\textbf{AutoML and Diversity.} AutoML frameworks utilize diversity primarily as a mechanism for error decorrelation in ensemble models. However, this performance-centric perspective overlooks the broader implications of model multiplicity. Recent studies have introduced the concept of xAutoML, aiming to make these pipelines interpretable. \citet{Zhai2024} proposed a domain-specific xAutoML framework using knowledge-informed feature extraction and model-agnostic selection methods. Similarly, \citet{Karthikeyan2025} combined the H2O AutoML system with LIME and SHAP methods, demonstrating the utility of hybrid systems by emphasizing that accuracy alone is insufficient for clinical trust. Despite these advances, most xAutoML tools still focus on explaining a single best model instead of accounting for the entire landscape of viable solutions.

\textbf{The Rashomon Effect.} The Rashomon effect \citet{breiman_rashomon_effect} describes the existence of multiple models that achieve similar predictive performance while offering different explanations. This phenomenon is closely related to model multiplicity and model under-specification. \citet{cashomon} extended this definition to CASHomon sets, considering multiple model classes and hyperparameters simultaneously. \citet{black2022} further refined this notion by distinguishing between procedural multiplicity, where models differ in their internal decision-making mechanisms, and predictive multiplicity, where they produce different predictions for the same observations. As noted by \citet{anders2020fairwashing}, procedural multiplicity is critical for detecting fairwashing scenarios, where model internals are manipulated without changing outputs.

Rashomon sets are characterized by global metrics such as the Rashomon Ratio \citep{rashomon_ratio_volume}, Pattern Rashomon Ratio \citep{semanova_noise}, Ambiguity \citep{marxpredictive}, and Discrepancy \citep{marxpredictive}, which define the overall size and level of disagreement within the set. Local metrics, including Viable Prediction Range (VPR) and Rashomon Capacity, provide instance-level insights into prediction uncertainty. To visualize these differences, variable importance clouds \citep{dongExploringCloudVariable2020} have been proposed to illustrate the range of feature importance across all models in a Rashomon set.

\textbf{Consequences in Rashomon Effect, Explainability, and AutoML.} The omission of the Rashomon perspective in AutoML processes leads to various consequences, both negative and positive. On the negative side, selecting a single model leads to model selection bias and the risk of x-hacking \citep{sharma2024x}, where users may selectively choose a model that supports a desired narrative. \citet{Rawal2026} demonstrated that this multiplicity can be exploited for adversarial fairwashing, where explanatory methods are misled to hide discriminatory features. 

On the positive side, exploring the Rashomon set enables uncertainty estimation \citep{cavus2025beyond, cavuspdp2026} and ensures that selected models align with ethical constraints. \citet{mullerEmpiricalEvaluationRashomon2023} provided an empirical evaluation showing that models within Rashomon sets exhibit high solution diversity, which can be measured through feature importance disagreement. Furthermore, \citet{Bifarin2024} argued that relying on a single best model in complex domains such as metabolomics can lead to misleading biological conclusions, and analyzing the entire solution landscape is essential. To formalize these trade-offs, the MIMOSA framework \citep{guidotti2025mimosa} integrates fairness, privacy, and causality into the generation of interpretable models, providing a theoretical foundation for the trustworthy AI analysis that ARSA ML aims to automate.

\section{ARSA ML Package}
\label{sec:arsa_package}

ARSA ML is an open source package publicly available at \url{https://pypi.org/project/arsa-ml/}. ARSA ML is built upon AutoGluon and H2O frameworks to generate diverse sets of trained models. Moreover, the package remains framework-agnostic, supporting the analysis of any externally trained models supplied in a compatible format (see Appendix~\ref{sec:arsa_package} for a full description of the package architecture). Figure~\ref{fig:package_overview} illustrates the general workflow where, as a result, users get an interactive Streamlit application, described in Appendix~\ref{sec:app_description}.

\begin{lstlisting}[language=Python, caption={ARSA ML code example}]
from arsa_ml.pipelines.builder_abstract import *
from arsa_ml.pipelines.pipelines_user_input import * 

#create pipeline from H2O saved models
builder = BuildRashomonH2O(models_directory=example_models_path, test_data = test_h2o, target_column=target_column, 
                           df_name = 'heart', base_metric='accuracy',  feature_imp_needed=True)

#preview Rashomon set properties
builder.preview_rashomon()

#set epsilon value
builder.set_epsilon(0.03)

#launch pipeline
rashomon_set, visualizer = builder.build()

#close dashboard 
builder.dashboard_close()

\end{lstlisting}

\subsection{Rashomon Analysis Metrics}
\label{sec:rashomon_metrics}
In ARSA ML we adopt and unify existing Rashomon analysis metrics for systematic analysis of AutoML pipelines.

Let $\mathcal{H}$ denote the hypothesis space of candidate models and let $M: \mathcal{H} \rightarrow \mathbb{R}$ denote an evaluation metric, where higher values indicate better performance. For a given metric $M$, we denote by $h_0^M \in \mathcal{H}$ the reference model, taken throughout to be the best-performing model returned by the AutoML system under metric $M$. All Rashomon set constructions and derived metrics are defined relative to this reference. Where a single metric is used and unambiguous, we write $h_0^M$ consistently; where two metrics $M_1$ and $M_2$ are considered simultaneously, the corresponding reference models $h_0^{M_1}$ and $h_0^{M_2}$ may differ. We denote by $n$ the number of observations in the dataset, by $x_i$ the $i$-th observation. 

We consider the notion of the Rashomon set, defined as the collection of models whose performance is within a tolerance level $\epsilon > 0$ of a reference model $h_0^M$:
\begin{equation}
R_{\epsilon}(h_0^M, M) = \{ h \in \mathcal{H} : M(h) \geq M(h_0^M) - \epsilon \}.
\end{equation}

To characterize the Rashomon set along complementary dimensions, we employ a hierarchy of metrics spanning set-level size, population-level predictive multiplicity, and instance-level diversity. 

\textbf{Set-level metrics.} 
The \emph{Rashomon Ratio} \citep{semanova_noise} measures the proportion of near-optimal models:
\begin{equation}
\hat{R}_{ratio}(\mathcal{H}, \epsilon) = |R_{\epsilon}(h_0^M, M)|\,\big/\,|\mathcal{H}|.
\end{equation}
A high Rashomon Ratio indicates that near-optimal performance is widespread across the hypothesis space, suggesting that architectural or algorithmic choices have limited impact on predictive quality alone.

However, two models may be architecturally distinct yet produce identical predictions on the observed data. To capture behaviorally meaningful diversity, rather than model count, we use the \emph{Pattern Rashomon Ratio} \citep{rashomon_ratio_volume, semanova_noise}:
\begin{equation}
\hat{R}_{ratio}^{pat}(\mathcal{H}, \epsilon) = |\pi(\mathcal{H}, \epsilon)|\,\big/\,|\psi(\mathcal{H})|,
\end{equation}
where $\pi(\mathcal{H}, \epsilon)$ and $\psi(\mathcal{H})$ denote unique prediction vectors for Rashomon-set models and all models in $\mathcal{H}$, respectively. These metrics quantify the number and diversity of near-optimal models.

\textbf{Population-level metrics.} We quantify disagreement among near-optimal models using \emph{ambiguity} and \emph{discrepancy} \citep{marxpredictive}, defined with respect to a fixed reference model $h_0^M$. For class predictions, ambiguity is defined as
\begin{equation}
\alpha_{\epsilon}(h_0^M) = \frac{1}{n} \sum_{i=1}^n \max_{h \in R_{\epsilon}(h_0^M, M)} \mathbbm{1}[h(x_i) \neq h_0^M(x_i)],
\end{equation}
and measures the fraction of observations for which at least one near-optimal model disagrees with the reference model. Discrepancy is defined as
\begin{equation}
D_{\epsilon}(h_0^M) = \max_{h \in R_{\epsilon}(h_0^M, M)} \frac{1}{n} \sum_{i=1}^n \mathbbm{1}[h(x_i) \neq h_0^M(x_i)],
\end{equation}
and captures the maximum proportion of predictions that may change when switching models. For probabilistic predictions, these metrics are extended using a threshold $\delta > 0$ \citep{probab} to count probability differences exceeding $\delta$. All definitions naturally extend to multiclass classification via $\arg\max$ \citep{capacity}.

\textbf{Instance-level metrics.} We further consider instance-level metrics, defined for each observation $x_i$. The \emph{Viable Prediction Range} (VPR) \citep{probab} captures the range of predicted probabilities across Rashomon models:
\begin{equation}
V_{\epsilon}(x_i) = \left[ \min_{h \in R_{\epsilon}} h(x_i)_1,\ \max_{h \in R_{\epsilon}} h(x_i)_1 \right],
\end{equation}
quantifying uncertainty in risk estimates for a given instance.

To measure predictive multiplicity at the instance level, we use \emph{Rashomon Capacity} \citep{capacity}. Let $\mathcal{M}_\epsilon(x_i) = \{h(x_i): h \in R_\epsilon(h_0^M, M)\}$ be the set of probability 
predictions generated by all models in the Rashomon set for observation $x_i$. 
The \emph{Rashomon Capacity} is defined as
\begin{equation}
    m_C(x_i) = 2^{C(\mathcal{M}_\epsilon(x_i))}, \quad
    C(\mathcal{M}_\epsilon(x_i)) = \sup_{P_Z} \inf_{q \in \Delta_c} 
    \mathbb{E}_{h \sim P_Z}\!\left[D_{KL}(h(x_i)\,\|\,q)\right],
\end{equation}
where $P_Z$ ranges over all probability distributions on $R_\epsilon(h_0^M, M)$, $\Delta_c$ denotes the probability simplex over $c$ classes, $q \in \Delta_c$ is a reference distribution, and $D_{KL}$ is the Kullback--Leibler divergence.

\textbf{Rashomon Intersection.} To account for multiple evaluation criteria, we introduce the \emph{Rashomon Intersection}, defined as the intersection of Rashomon sets constructed for different metrics:
\begin{equation}
RI_{\epsilon}(M_1, M_2) = R_{\epsilon}(h_0^{M_1}, M_1) \cap R_{\epsilon}(h_0^{M_2}, M_2).
\end{equation}
It captures models that are simultaneously near-optimal with respect to both $M_1$ and $M_2$. Since the reference model may differ across metrics, any downstream analysis requires a single joint reference model. We define it as the solution to the weighted optimization problem over the intersection:
\begin{equation}
h_0 = \arg\max_{h \in RI_{\epsilon}(M_1, M_2)} \bigl( w_1 M_1(h) + w_2 M_2(h) \bigr), \quad w_1, w_2 \in [0,1],\; w_1 + w_2 = 1.
\end{equation}
 
 We propose three methods for determining $w_1$ and $w_2$: (1) \emph{Custom weights} The user specifies $w_1$ and $w_2$ directly, reflecting domain-specific priorities. (2) \emph{Entropy method} \citep{Wang2023}. Weights are derived from the entropy of each metric's distribution across models within Rashomon Intersection. (3)\emph{CRITIC method} \citep{Wang2023}. This is an objective weighting technique used in multi-criteria decision-making. Details of these methods can be found in Appendix~\ref{sec:weight_intersection}.

\section{Explanation Hackability}
\label{sec:xhack}
 
The metrics introduced in Section~\ref{sec:rashomon_metrics} characterize the Rashomon effect along predictive dimensions: how many near-optimal models exist, how much their predictions differ across the population, and how uncertain the predicted probability of a single observation is. However, they do not address a complementary question of practical importance: \emph{do near-optimal models agree on which features drive the prediction?}

If the Rashomon set contains near-optimal models that assign substantially different importance to the input features, then the choice of which model to report determines which explanation is delivered --- without any sacrifice in predictive performance. We formalise this risk as \emph{explanation hackability} (XHack): the degree to which a framework's Rashomon set permits the selection of models that support conflicting feature-importance narratives at no performance cost.

\textbf{Feature importance metric.} Let $\F = \{f_1, \ldots, f_p\}$ denote the set of input features. Each model $h \in \mathcal{H}$ is equipped with a feature importance ranking $\varphi(h) : \F \to \{1, \ldots, p\}$, where $\varphi(h)(f) = 1$ denotes the most important feature according to $h$. For a feature $f \in \F$ and a threshold $k \in \{1, \ldots, p\}$, define the \emph{top-$k$ importance metric}:
\begin{equation}
    \phifk(h) =\mathbbm{1}\left[\,\varphi(h)(f) \leq k\,\right] \in \{0,\, 1\},
    \label{eq:phi}
\end{equation}
which equals $1$ if model $h$ considers $f$ among its $k$ most important features, and $0$ otherwise.

\textbf{Rashomon Intersection for feature importance.} We instantiate the Rashomon Intersection from Section~\ref{sec:rashomon_metrics} with $M_1 = M$ (a predictive performance metric) and $M_2 = \phifk$ (the top-$k$ importance indicator for feature $f$) and since $\phifk$ is binary, its Rashomon set at any $\epsilon < 1$ reduces exactly to models that rank $f$ in their top-$k$ :
\begin{equation}
    \RI{\epsilon}(M,\, \phi^{f,k}) = R_{\epsilon}(h_0^M,\, M) \cap R_{\epsilon}(h_0^{\phifk},\, \phifk) = \left\{\, h \in R_{\epsilon}(h_0^M,\, M) \;\big|\; \varphi(h)(f) \leq k \, \right\}.
    \label{eq:ri_feat}
\end{equation}

This is the set of near-optimal models that also consider $f$ an important feature. A large $|\RI{\epsilon}(M, \phifk)|$ relative to $|R_{\epsilon}(h_0^M, M)|$ means that performance-competitive models consistently agree on the importance of $f$; a small intersection means that $f$'s importance is specific to only a minority of near-optimal models.

\textbf{XHackability.} The intersection $\RI{\epsilon}(M, \phifk)$ characterizes the explanation stability of a specific feature $f$. To obtain a single, comparable scalar that captures the overall hackability of a framework's explanation landscape, we ask: across all features, what is the largest fraction of near-optimal models that agree on any single feature's importance? XHackbility is the complement of this maximum:
 
\begin{definition}[Explanation Hackability]
\label{def:xhack}
The \emph{explanation hackability} for given dataset with feature set $F$, performance metric $M$, tolerance $\epsilon$, and top-$k$ threshold is:
\begin{equation}
    \xhack(k,\, M,\, \epsilon) = 1 - \max_{f\, \in\, \F}\; \frac{\left|\,\RI{\epsilon}(M,\, \phifk)\,\right| }{\left|\,R_{\epsilon}(h_0^M,\, M)\,\right|}.
    \label{eq:xhack}
\end{equation}
\end{definition}
 
\noindent where the term $\max_{f \in \F} |\RI{\epsilon}(M, \phifk)| / |R_\epsilon(h_0^M, M)|$ identifies the feature whose importance is most consistently shared across near-optimal models; XHack measures how far even this best-case feature falls from universal agreement. $\xhack \in [0,1]$: at $\xhack = 0$, every near-optimal model agrees on at least one feature's top-$k$ membership, so no narrative manipulation is possible within $R_\epsilon$ without sacrificing performance; at $\xhack = 1$, no feature is consistently ranked in the top-$k$, meaning any desired feature-importance narrative can be supported by some near-optimal model. Intermediate values scale accordingly --- for instance, $\xhack = 0.6$ implies that even the most consistently important feature appears in the top-$k$ of only $40\%$ of near-optimal models.

\section{Benchmark of Rashomon set Diversity in AutoGluon and H2O}
\label{sec:results}

\subsection{Methodology}

\textbf{Datasets.}
Experiments are conducted on 30 binary classification datasets retrieved from OpenML~\citep{bischl2017openml}; these datasets are widely used in benchmarks like TabArena~\citep{tabarena}. Finally, only 28 are reported here since technical difficulties with training of AutoML frameworks for two of 30. Only binary target datasets are included. The default target column defined by OpenML is used as the response variable in each case. Before training, the target variable is integer-encoded. For evaluation, we use stratified 4-fold cross-validation. 

\textbf{Model Training.}
Since ARSA\;ML package, AutoML frameworks are used for model training: AutoGluon and H2O AutoML. AutoGluon trains models with the \textit{good\_quality} preset and a time budget of 2h per fold. H2O AutoML is configured with an equivalent time limit and capped at 20 models. This limitation is introduced due to out-of-memory errors encountered during training on the server machine, which caused instability when a larger number of models was allowed. Both frameworks handle internal preprocessing automatically, including missing value imputation, categorical encoding, and feature normalization.

\textbf{Feature Importance.} We extract feature importance rankings via native APIs, applying targeted adjustments to ensure cross-framework consistency at the original-feature level. In AutoGluon, we compute permutation importance uniformly across all models using the framework’s training set evaluation. While we utilize H2O’s default permutation importance for most models, we compute it post-hoc for stacked ensembles and utilize weight-based importance for Deep Learning models. To reconcile H2O’s internal feature discretization with our required granularity, we assign each original feature the minimum rank of its constituent bins and recompute the final rankings across the original feature set.

\textbf{Rashomon set Construction.}
All computations are performed using ARSA ML and experimental code is available on GitHub\footnote{\url{https://github.com/sienkozuzanna/ARSA_experiments/}}.  Six evaluation metrics are considered: accuracy, balanced accuracy, F1, precision, recall, and ROC AUC. For each metric, seven epsilon thresholds are applied, defined as fixed percentages of the best observed score: $1\%$, $2.5\%$, $5\%$, $10\%$, $15\%$, $20\%$, and $25\%$. This yielded $42$ Rashomon set configurations per fold (6 metrics $\times$ 7 epsilon levels). Results are stored in json files available in Zendo\footnote{\url{https://doi.org/10.5281/zenodo.20191299}}.

\subsection{Results}

Table~\ref{tab:base_performance} reports the reference model performance across 28 benchmark datasets under six evaluation metrics. The results reveal a clear split between frameworks: AutoGluon achieves higher accuracy, precision, and ROC AUC, winning the majority of datasets under these metrics, while H2O dominates on balanced accuracy, F1, and recall, where it wins 21, 22, and 25 datasets, respectively.

\begin{table}[ht]
  \centering
  \caption{Reference model performance across 28 benchmark datasets.}
  \label{tab:base_performance}
  \footnotesize
  \begin{tabular}{l cc cc r}
    \toprule
    Metric & \multicolumn{1}{c}{AG} & \multicolumn{1}{c}{H2O} & W\textsubscript{AG} & W\textsubscript{H2O}  & $p$ \\
    \midrule
    Accuracy & \textbf{0.8827 $\pm$ 0.0857} & 0.8751 $\pm$ 0.0887 & 24 & 4  & $<$0.001 \\
    Balanced Accuracy & 0.7189 $\pm$ 0.1382 & \textbf{0.7641 $\pm$ 0.1021} & 7 & 21  & $<$0.001 \\
    F1 & 0.5943 $\pm$ 0.2928 & \textbf{0.6378 $\pm$ 0.2338} & 5 & 22  & $<$0.001 \\
    Precision & \textbf{0.7926 $\pm$ 0.1441} & 0.6581 $\pm$ 0.2359 & 22 & 5  & $<$0.001 \\
    Recall & 0.5593 $\pm$ 0.3185 & \textbf{0.7407 $\pm$ 0.1864} & 2 & 25 &  $<$0.001 \\
    ROC AUC & \textbf{0.8557 $\pm$ 0.0908} & 0.8523 $\pm$ 0.0902 & 18 & 10  & 0.009 \\
    \bottomrule
  \end{tabular}
  \begin{tablenotes}
    \small
    \item Mean $\pm$ std computed across 28 datasets (scores averaged over folds per dataset). W = number of datasets won. $p$: paired Wilcoxon signed-rank test (two-sided). Bold indicates the better mean.
  \end{tablenotes}
\end{table}

\begin{figure}
    \centering
    \begin{subfigure}{0.3\textwidth}
        \includegraphics[width=\linewidth]{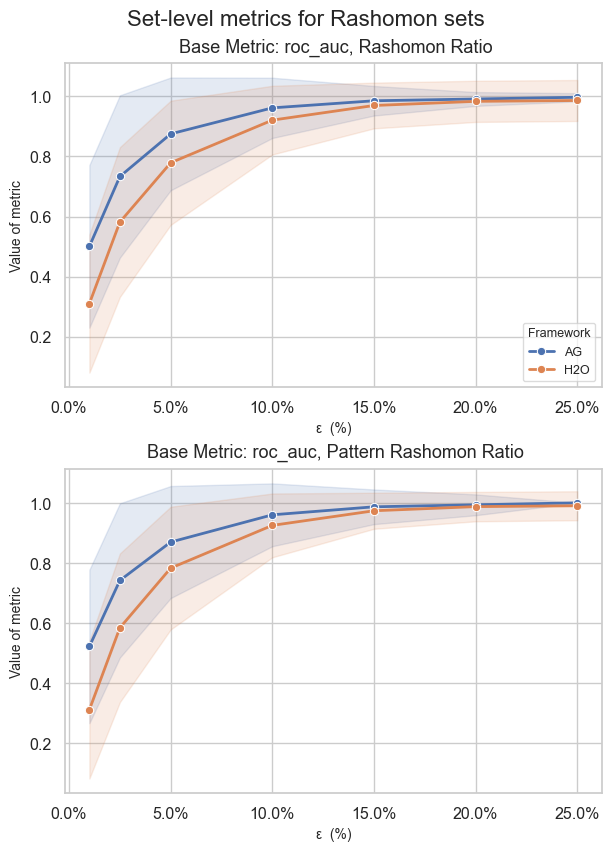}
        \caption{Set-level metrics for range of epsilons}
        \label{fig:set_metrics_auc}
    \end{subfigure}
    \hfill
        \begin{subfigure}{0.3\textwidth}
            \includegraphics[width=\linewidth]{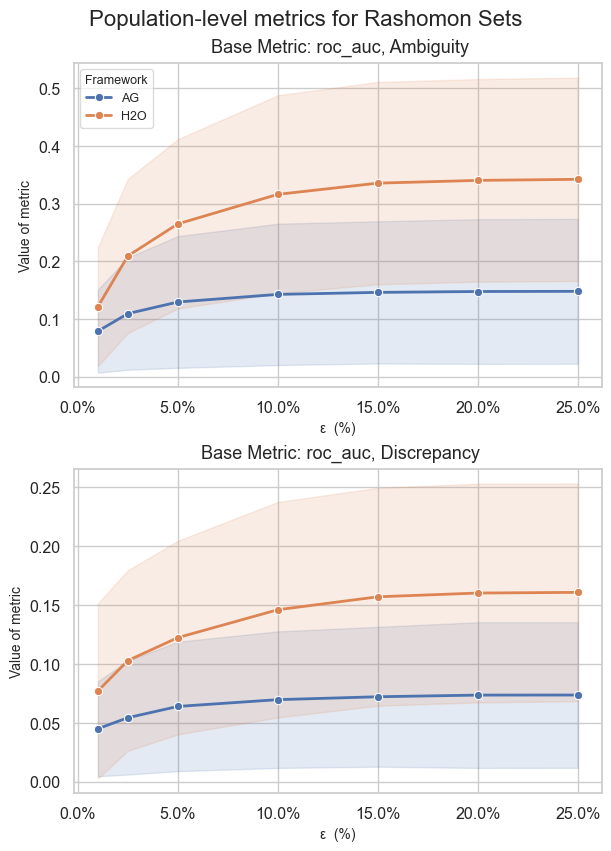}
            \caption{Population-level metrics for range of epsilons}
            \label{fig:pop_metrics_auc}
        \end{subfigure}
    \hfill
        \begin{subfigure}{0.3\textwidth}
        \includegraphics[width=\linewidth]{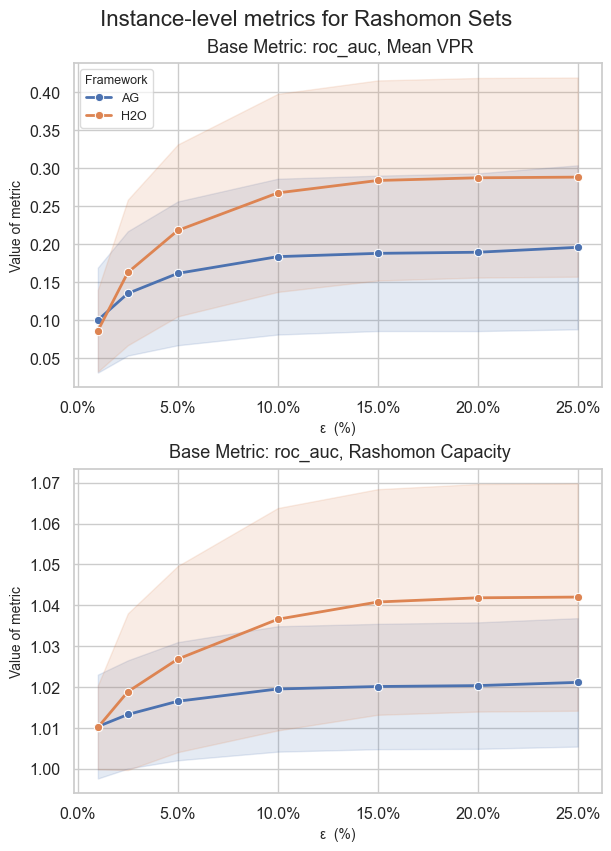}
                \caption{Instance-level metrics for range of epsilons}
        \label{fig:inst_metrics_auc}
    \end{subfigure}
\end{figure}

\textbf{RQ1: Rashomon set size and composition.} Figure~\ref{fig:set_metrics_auc} shows how the Rashomon set evolves with increasing $\epsilon$ under the ROC AUC metric. As $\epsilon$ increases, all metrics grow and then stabilize, indicating that a larger tolerance expands the set but with diminishing gains in diversity. AutoGluon consistently exhibits a higher Rashomon Ratio, suggesting a larger set of near-optimal models. However, the Pattern Rashomon Ratio is similar across frameworks, implying that many of these models produce similar predictions. In contrast, H2O achieves comparable behavioral diversity with fewer models.

\begin{figure}
        \centering
        \includegraphics[width=\linewidth]{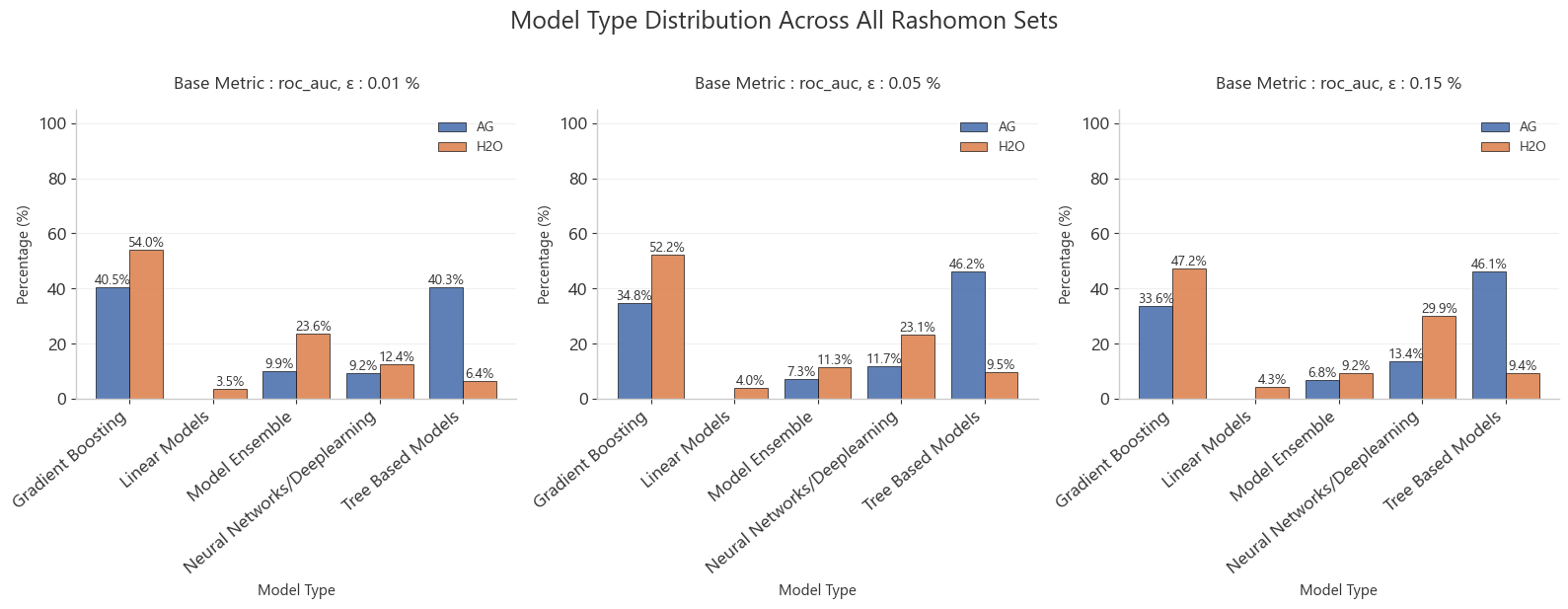}
        \caption{Differences in algorithm types within Rashomon sets between frameworks.}
        \label{fig:auc_epsilon_05_model_types}
\end{figure}

Figure~\ref{fig:auc_epsilon_05_model_types} presents the distribution of model types within the Rashomon Sets at three $\epsilon$ levels under the ROC AUC metric. The two frameworks exhibit markedly different compositional profiles. AutoGluon's Rashomon Sets are dominated by Tree-Based Models and Gradient Boosting models, with this distribution remaining stable across all $\epsilon$ levels, suggesting uniform representation of algorithm families across performance tiers. Notably, AutoGluon's Gradient Boosting category encompasses multiple distinct algorithms -- LightGBM, XGBoost, and CatBoost -- and similarly, its Tree-Based Models comprise both Random Forest and Extra Trees variants, making AutoGluon's effective algorithmic diversity even broader than the aggregated bars suggest. H2O, by contrast, is heavily concentrated on Gradient Boosting, represented primarily by GBM, while the share of Neural Networks grows substantially with $\epsilon$ -- from 12.4\% at $\epsilon = 0.01$ to 29.9\% at $\epsilon = 0.15$ -- indicating that H2O's Deep Learning models enter the set only as the tolerance widens.

\textbf{RQ2: Predictive multiplicity at population and instance level.} Figure~\ref{fig:pop_metrics_auc}  H2O shows higher ambiguity and discrepancy across all $\epsilon$ levels, indicating stronger predictive multiplicity and less consistent predictions. This pattern is reinforced by higher VPR and slightly higher Rashomon Capacity shown in Figure~\ref{fig:inst_metrics_auc}, suggesting greater instance-level uncertainty. Overall, AutoGluon produces larger but more homogeneous Rashomon sets, while H2O yields smaller yet more diverse and uncertain prediction behaviors.

\begin{figure}
    \centering
    \includegraphics[width=0.7\linewidth]{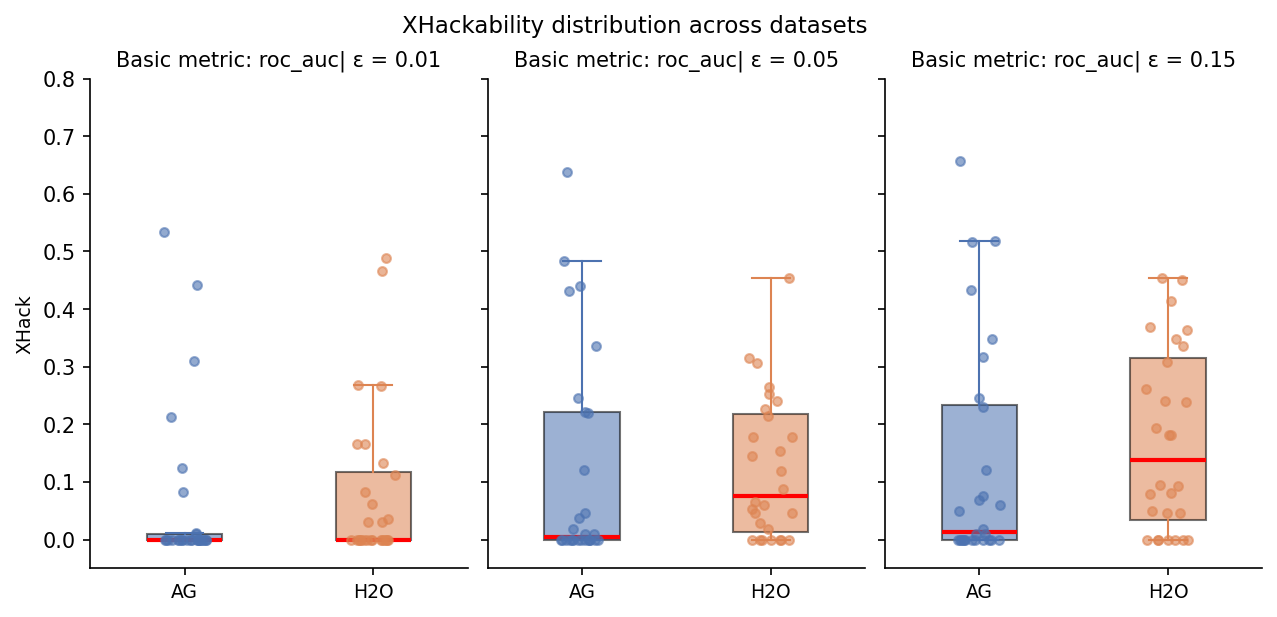}
    \caption{Distribution of XHackability scores across 28 datasets at increasing $\epsilon$ levels (ROC AUC metric). Each box shows the interquartile range across datasets; the line is the median. AutoGluon (left) shows medians close to zero that grow slowly with $\epsilon$; H2O (right) exhibits higher and more dispersed values throughout, indicating greater susceptibility to explanation manipulation.}
    \label{fig:xhack}
\end{figure}

\textbf{RQ3: Explanation hackability.} Figure~\ref{fig:xhack} shows the distribution of $\xhack$ scores across 28 datasets at three $\epsilon$ levels under ROC AUC. AutoGluon explanations are substantially more consistent: for most datasets, the  $\xhack$ is close to zero at small $\epsilon$, meaning that near-optimal AutoGluon models largely agree on which features matter most. H2O, by contrast, exhibits markedly higher and more variable $\xhack$ values across the same datasets, indicating that its Rashomon sets routinely contain models supporting conflicting feature-importance narratives at no performance cost. In both frameworks $\xhack$ increases with $\epsilon$, as expected: a wider tolerance admits more models and therefore more explanation diversity. However, the gap between frameworks persists across all $\epsilon$ levels, suggesting a structural difference rather than an artefact of threshold choice. This finding is consistent with the set-composition results in Figure~\ref{fig:auc_epsilon_05_model_types}: H2O's Rashomon sets are dominated by GBM variants whose feature rankings can diverge substantially, while AutoGluon's broader mix of tree ensembles converges on more stable importance orderings. Together, these results imply that H2O users face a considerably higher risk of x-hacking.

\section{Conclusions}

AutoML systems have made high-performing predictive models widely accessible, yet they typically return a single "best" model while ignoring many equally valid alternatives. ARSA ML addresses this gap by making these alternatives visible and measurable. By converting a broad set of Rashomon metrics into an automated, framework-agnostic pipeline, it allows practitioners to quantify the predictive and explanatory flexibility within their model set before committing to a final deployment.

Our benchmark of AutoGluon and H2O reveals a consistent asymmetry in how this flexibility manifests: the two frameworks occupy opposite ends of a size–diversity–hackability trade-off. AutoGluon generates larger, algorithmically diverse sets that remain prediction-homogeneous with stable explanations. In contrast, H2O generates compact sets concentrated in a single model family, yet these sets exhibit greater prediction divergence and explanation instability. Because neither framework currently exposes these dynamics, ARSA ML is essential for making this underlying behavior actionable.

These technical trade-offs connect directly to an emerging regulatory crisis. As \citet{frohnapfel2026} argues, predictive multiplicity conflicts with the EU AI Act’s requirement that high-risk systems report accuracy not just at the dataset level, but for specific individuals. If two near-optimal models disagree on a single person’s outcome, the choice of one model over another becomes an arbitrary decision with significant ethical consequences. By operationalizing metrics like ambiguity, discrepancy, Viable Prediction Range, and Rashomon Capacity, ARSA ML integrates compliance-relevant data directly into the AutoML pipeline, making legal and statistical accountability accessible to every practitioner.

\begin{acknowledgements}
This research was carried out with the support of the High Performance Computing Center at Faculty of Mathematics and Information Science Warsaw University of Technology.
\end{acknowledgements}


\bibliography{references}

\newpage
\appendix

\section{ARSA\;ML Package description}
\begin{figure}[!h]
    \centering
    \includegraphics[width=0.95\linewidth]{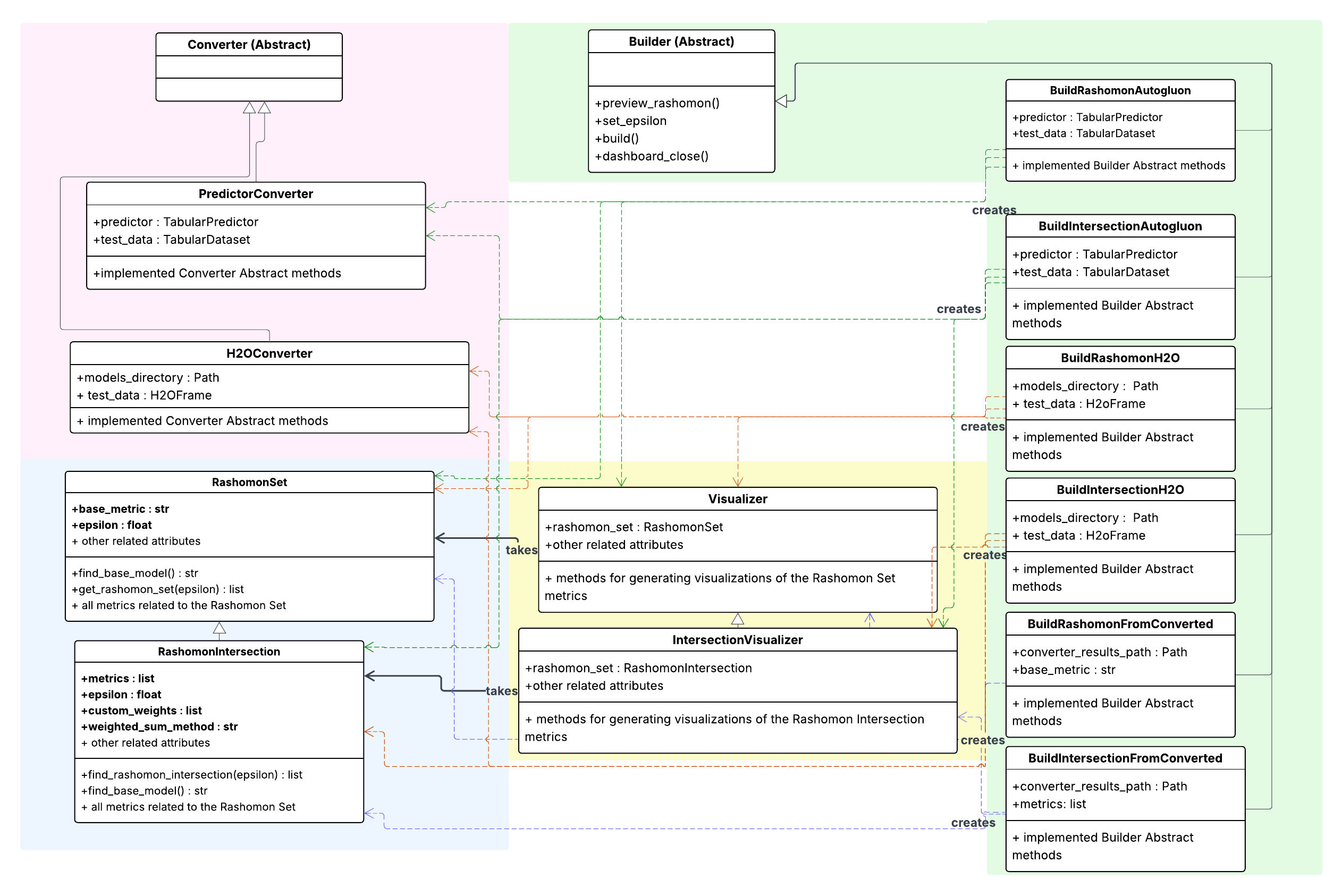}
    \caption{Technical schema of ARSA ML modules}
    \label{fig:pacakge_detailed_schema}
\end{figure}

\textbf{Converters.}
The Converters module normalizes the heterogeneous outputs of different AutoML frameworks into a unified internal representation: a model leaderboard, class-prediction vectors, probability-prediction matrices, and feature-importance rankings. Three concrete implementations are provided: \texttt{PredictorConverter} for AutoGluon, \texttt{H2OConverter} for H2O, and a base \texttt{Converter} interface for custom integrations. Each converter exposes a \texttt{convert()} method whose output is passed directly to the Rashomon Analysis module, and a \texttt{save\_results()} method for caching converted artefacts to disk.
 
\textbf{Rashomon Analysis.}
The core module implements two classes. \texttt{RashomonSet} accepts converter output together with a base metric $M$ and a tolerance $\epsilon$, constructs the set $R_{\epsilon}(h_0^M, M)$, and exposes methods for every metric describe in Section~\ref{sec:rashomon_metrics}: Rashomon Ratio, Pattern Rashomon Ratio, Ambiguity, Discrepancy (binary, probabilistic, and multiclass variants), Viable Prediction Range, Rashomon Capacity, Percent Agreemen Agreement Rate, and Cohen's Kappa. A \texttt{summarize\_rashomon()} met returns a consolidated view of all metrics for rapid inspection. \texttt{RashomonIntersection} extends \texttt{RashomonSet} to the multi-metric setting: it computes two independent Rashomon sets under metrics $M_1$ and $M_2$, takes their intersection, and selects a joint reference model via the weighted-sum approach described in Section~\ref{sec:rashomon_metrics} (custom weights, Entropy, or CRITIC). All single-metric metrics are inherited and recomputed on the intersection.
 
\textbf{Visualizers.}
The Visualizer module provides interactive Plotly-based plots covering the full set of Rashomon metrics:
\begin{itemize}
    \item gauge and scatter plots for Rashomon Ratio and Pattern Rashomon Ratio across $\epsilon$ values; 
    \item lollipop charts for Ambiguity and Discrepancy; 
    \item histograms and box plots for Rashomon Capacity and VPR widths;
    \item agreement bar charts and Cohen's Kappa heatmaps;
    \item a feature-importance heatmap across all models in the set.
\end{itemize} 

\texttt{IntersectionVisualizer} adds a Venn diagram of the two constituent Rashomon sets and a scatter plot of model scores with the Pareto front highlighted, enabling comparison with standard multi-objective optimization. All plots include hover tooltips for precise numerical inspection.
 
\textbf{Pipelines.}
The Pipelines module integrates the three preceding modules into end-to-end workflows that require minimal user effort. Seven concrete Pipeline classes cover every combination of input type (raw AutoGluon predictor, saved H2O models, pre-converted results) and analysis target (\texttt{RashomonSet} or \texttt{RashomonIntersection}). Each exposes three methods: \texttt{preview\_rashomon()}, which plots Rashomon set size against $\epsilon$ to guide threshold selection; \texttt{set\_epsilon()}; and \texttt{build()}, which constructs the analysis objects and launches a local Streamlit dashboard. This design allows practitioners to obtain a full Rashomon analysis from a trained AutoML predictor in three lines of code, while retaining access to individual modules for custom workflows.

\section{Streamlit application}
\label{sec:app_description}
An additional product of this work is a Streamlit web application published on the Streamlit Cloud, allowing users to experiment with the package functionalities in a no-code manner. It consists of four pages - Home page, Datasets page, Rashomon page, and the Intersection page. The initial view appearing after visiting the website provides a brief overview of the application's main purpose and key features, along with the intuitive explanations of the predictive multiplicity problem, the Rashomon Effect, and the introduced approach of the Rashomon Intersection. On the Datasets page, users can find detailed descriptions of the eight pre-saved datasets available for analysis. Finally, on the Rashomon page and on the Intersection page, users can construct the corresponding objects by selecting a dataset and specifying the related parameters using interactive widgets in order to view the dashboard for analysis.
\\

\begin{figure}[!ht]
\centering
\includegraphics[width=0.9\linewidth, keepaspectratio]{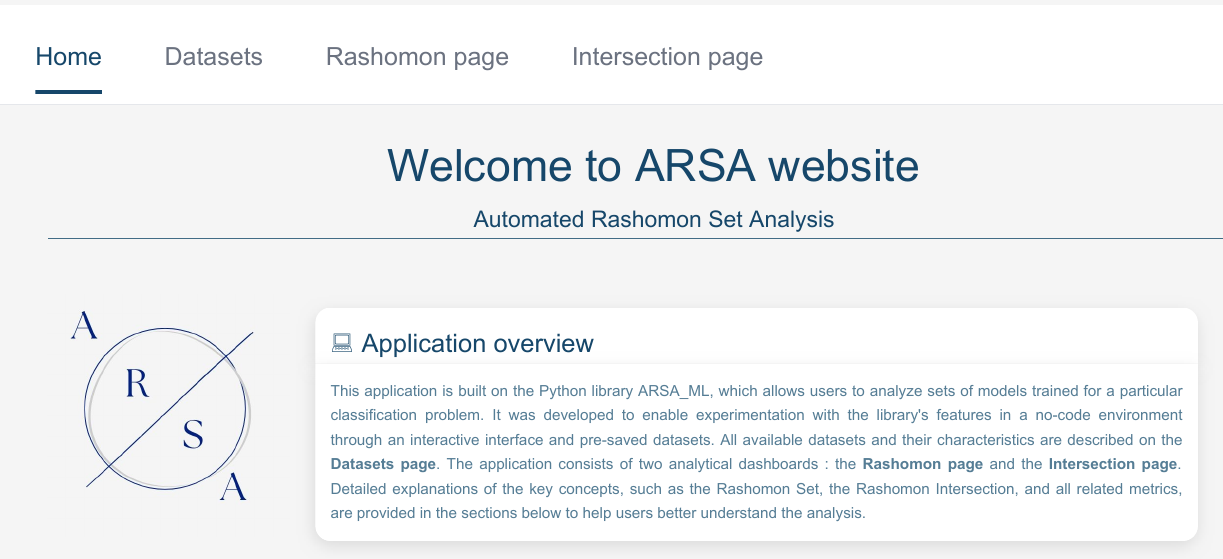} 
\caption{Initial screen of the ARSA ML web application. It contains a top navigation bar that allows switching between pages, a package logo, and a brief overview of the application's functionalities. Additional Home page elements are visible upon scrolling. }
\label{fig:homepage1}
\end{figure}

\begin{figure}[!ht]
    \centering
    \includegraphics[width=0.9\linewidth, keepaspectratio]{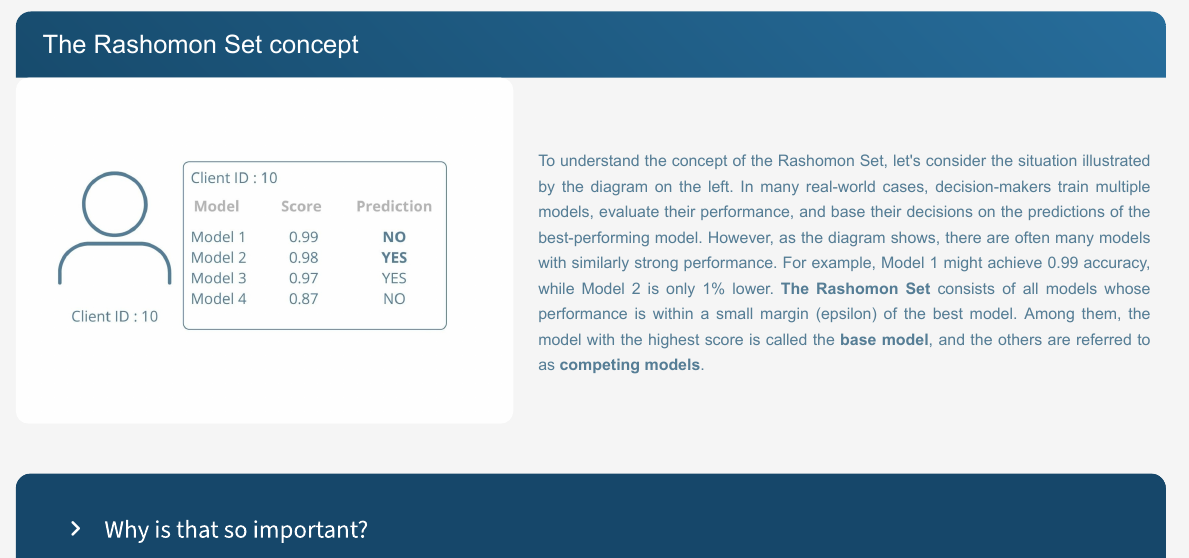} 
    \caption{One of the Home page elements - an intuitive explanation of the Rashomon set concept based on a sample scenario presented on a graphic. An expandable panel contains additional information about the importance of the detailed analysis of the predictive multiplicity problem before the decision-making process. }
    \label{fig:homepage2}
\end{figure}

\textbf{Home page}
\\
The home page contains the application overview and an intuitive explanation of the predictive multiplicity problem and the Rashomon Effect. On this page, users can gain an understanding of the key definitions and metrics illustrated on the dashboards, as well as access articles from the bibliography. Figure \ref{fig:homepage1} presents the application's Home screen that appears after visiting the website, while Figure \ref{fig:homepage2} illustrates an example section with the Rashomon set description. This page contains similar, intuitive explanations of the newly introduced concept of the Rashomon Intersection, along with all metrics related to the predictive multiplicity problem. We provided some real-life examples, along with diagrams, to allow easier understanding of these concepts without the wide knowledge of this topic. We decided not to include formal definitions, which can be found in the literature, in the application, as it would not be helpful for most of the users and would reduce the readability of the page. Instead, we provided links to the key articles used in this project.

\newpage

\textbf{Datasets page}
    \\
This page is dedicated to a detailed description of the pre-saved datasets available for experimentation. It contains information about the source, license, classification task type, and dataset characteristics supported by additional charts. For the purpose of this project, we selected eight datasets and divided them into three categories: 'business application', 'predictive multiplicity problem', and 'datasets challenging for ML algorithms'. The first category represents datasets that are related to employment, finance, and banking, such as a credit score classification dataset. The second category consists of datasets for which we find the predictive multiplicity to be especially problematic. Those datasets are related to recidivism, bias, or disease prediction. The last group contains datasets with class imbalance, feature correlation, or multiple classes, which are often causes of incorrect predictions produced by ML models. Table \ref{tab:datasets} presents information about the selected datasets.

\begin{table}[!ht]
\caption[Pre-saved datasets overview]{A short overview of the eight datasets selected as pre-saved datasets for analysis in the Streamlit application. Column 'Classes' contains the number of unique labels in the target column, 'Features' and 'Observations' inform about the number of features and samples in the dataset, and 'Category' reflects one of the $3$ categories described previously.}
\label{tab:datasets}
\centering
\begin{tabular}{|p{0.25\textwidth}|p{0.08\textwidth}|p{0.1\textwidth}|p{0.15\textwidth}|p{0.35\textwidth}|}
\hline
Dataset name & Classes & Features & Observations & Category \\ \hline
HR Job Change & $2$&  $13$ & $19 158$ & Business Application \\
Credit Score & $3$ & $7$ & $164$ & Business Application \\
Breast Cancer & $2$ & $32$ & $569$ & Predictive Multiplicity Problem \\
Heart Failure & $2$ & $11$ & $918$ & Predictive Multiplicity Problem \\
COMPAS & $2$ & $11$ & $6172$ & Predictive Multiplicity Problem \\
Glass Types & $6$ & $9$ & $214$ & Challenging For ML Algorithms \\
Letter Recognition & $26$ & $16$ & $20 000$ & Challenging For ML Algorithms \\
Yeast & $10$ & $9$ & $1484$ & Challenging For ML Algorithms\\
 \hline
\end{tabular}
\end{table}

\begin{figure}[!ht]
\centering
\includegraphics[width=0.9\textwidth]{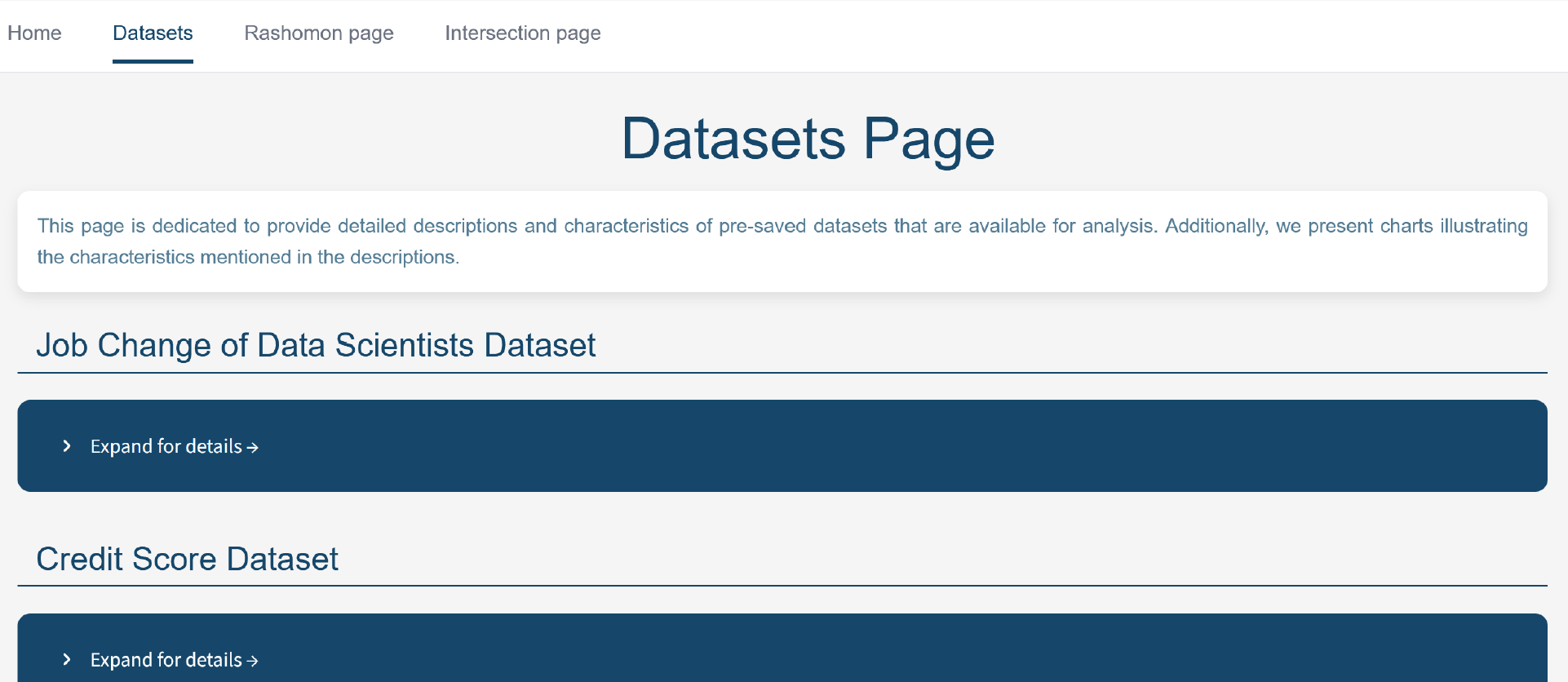}
\caption{The initial screen of the Datasets page with expandable panels allowing the exploration of every pre-saved dataset.}
\label{fig:datasets_page}
\end{figure}

\begin{figure}[!ht]
    \centering
    \includegraphics[width=0.9\textwidth, keepaspectratio]{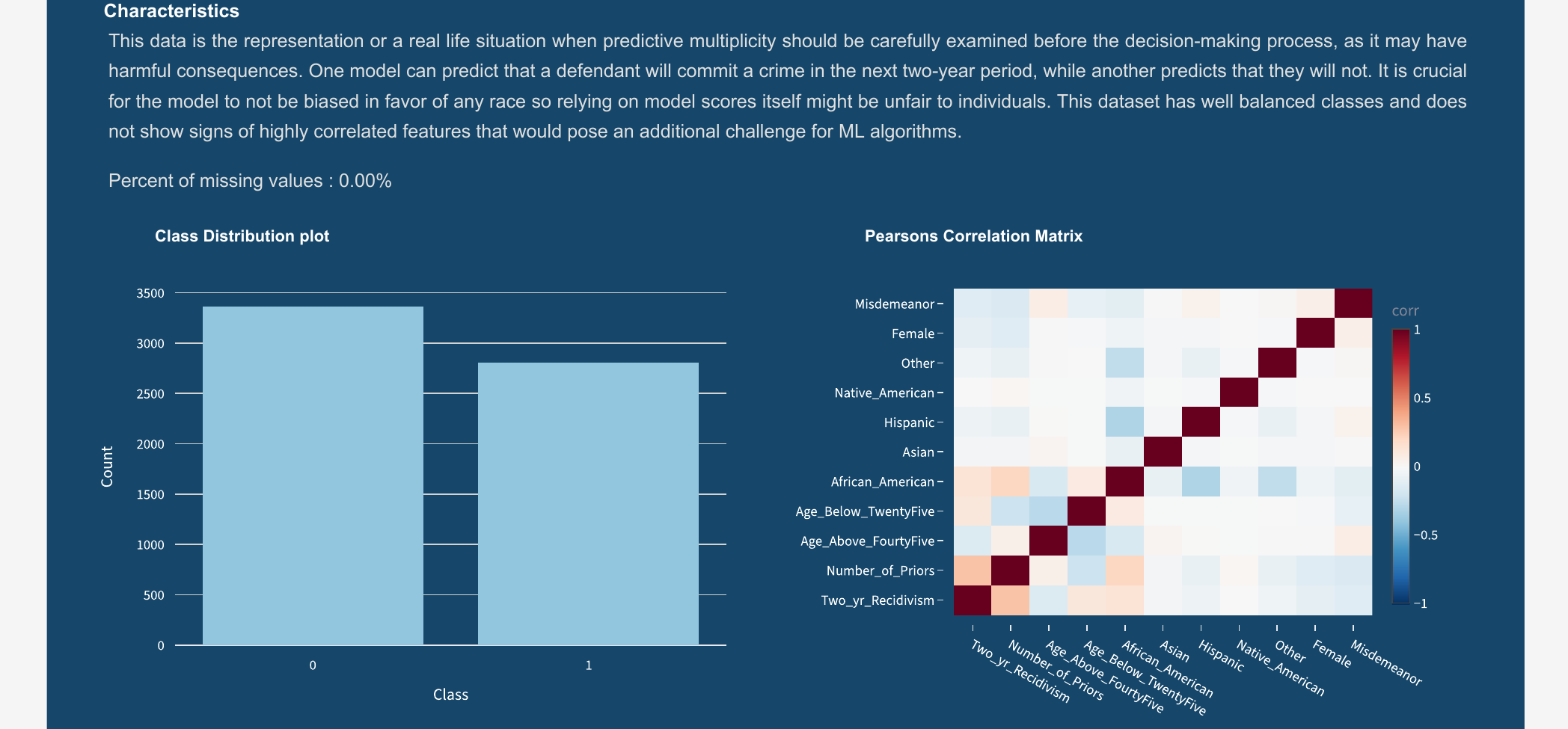} 
    \caption{A part of the expanded panel for the COMPAS dataset. It contains a short description of the dataset characteristics, the percentage of missing values found in a dataset, along with class distribution and numerical features correlation charts.}
    \label{fig:datasets_page2}
\end{figure}

\newpage
All datasets presented in Table \ref{tab:datasets} are described on the Datasets page. Figure \ref{fig:datasets_page} illustrates the view of the Datasets page before selecting a particular dataset for analysis. After choosing a dataset, users can expand a corresponding panel and view all information about the chosen dataset, such as the source link, license, feature description, and the dataset's characteristics. Figure \ref{fig:datasets_page2} presents a part of the panel's content for the COMPAS dataset containing its characteristics and the supporting charts.

\begin{figure}[!ht]
\centering
\includegraphics[width=0.9\textwidth]{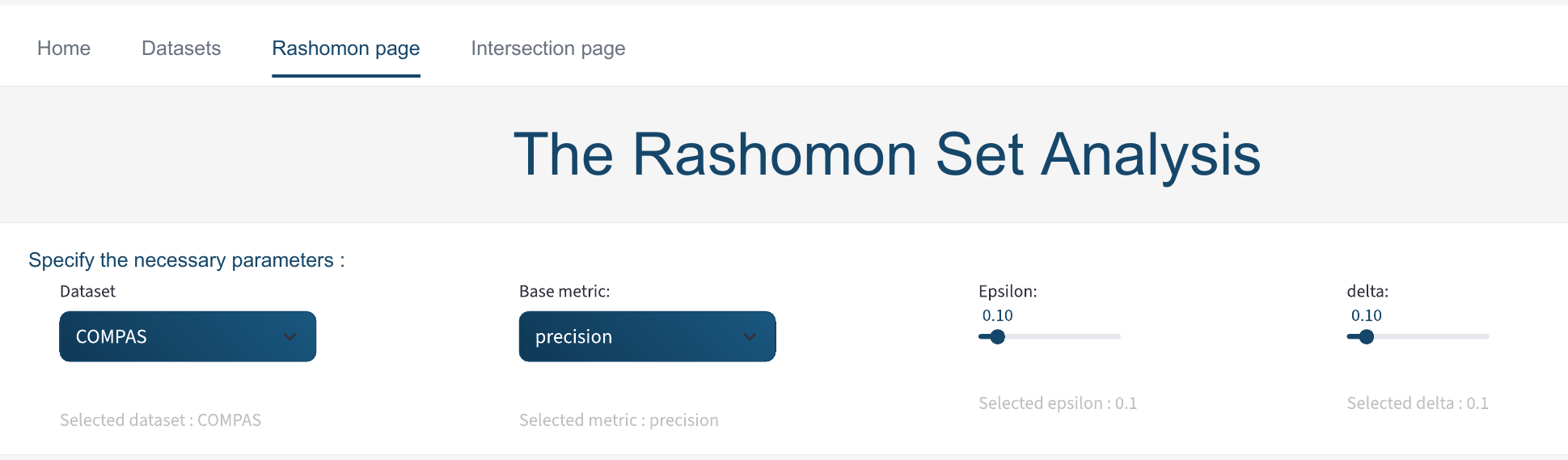} 
\caption{The initial view of the Rashomon page providing parameter selection using interactive widgets. After specifying each parameter, corresponding information appears below the widget.}
\label{fig:rashomon_widgets}
\end{figure}

\textbf{Rashomon page}
\\
The primary page of the application is the Rashomon page, where users can select custom parameters and analyze the properties of the created Rashomon set. Firstly, one of the pre-saved datasets should be selected for analysis. Then, widgets allowing the specification of other parameters, such as the base metric, epsilon, and delta (only for binary classification tasks), appear on the screen. The available widgets with sample parameter selection are illustrated in Figure \ref{fig:rashomon_widgets}.

\begin{figure}[!ht]
\centering
\includegraphics[width=0.9\linewidth, keepaspectratio]{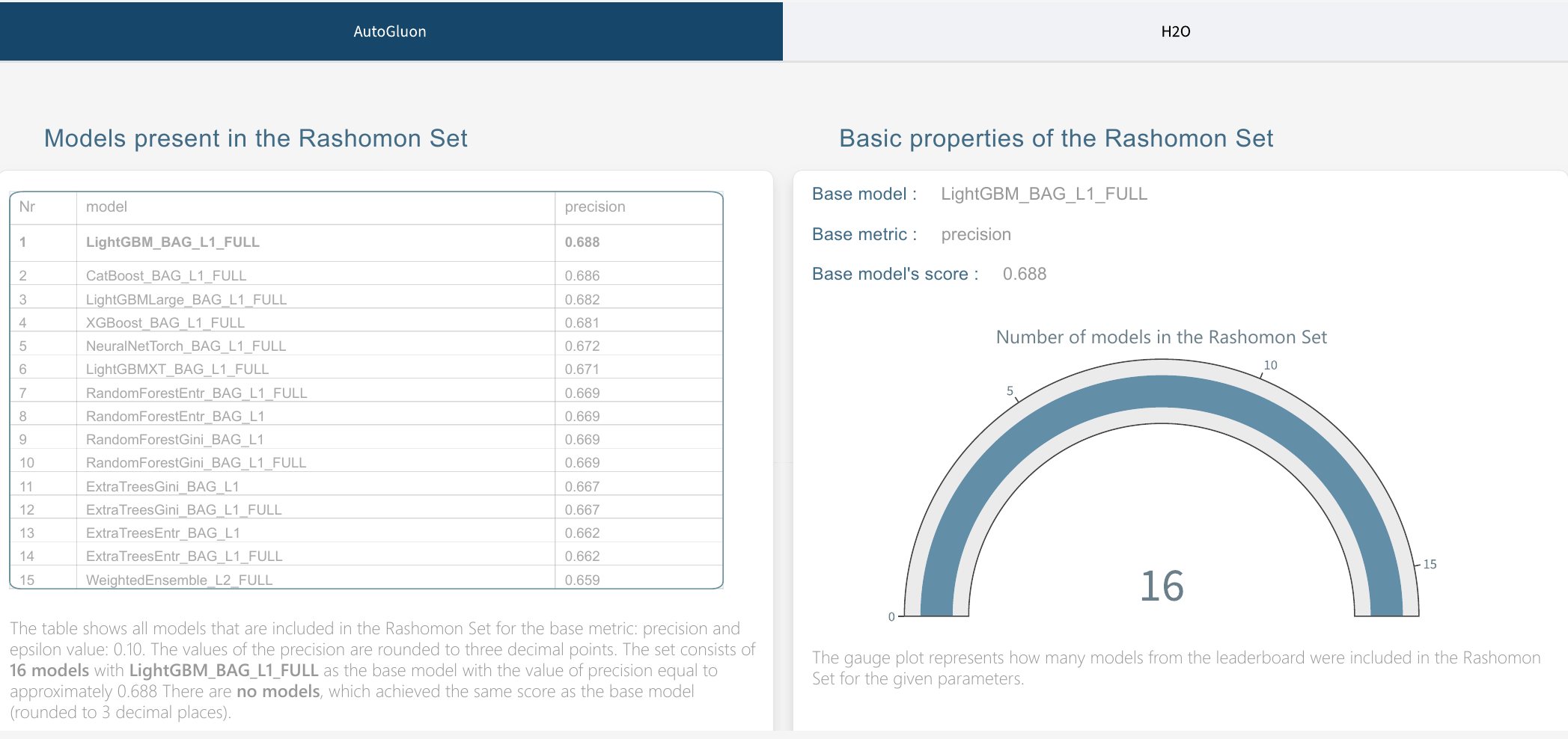} 
\caption{Initial screen of the Rashomon set analysis dashboard. On the left-hand side of the dashboard, users may find a table containing all models that are included in the Rashomon set for specified parameters, as well as a brief description of the table properties. On the right-hand side, there is a gauge plot illustrating the fraction of models from the leaderboard that are a part of the Rashomon set, along with the selected parameters and the name of the reference model.}
\label{fig:rashomon1}
\end{figure}

\begin{figure}[!ht]
    \centering
    \includegraphics[width=0.9\linewidth, keepaspectratio]{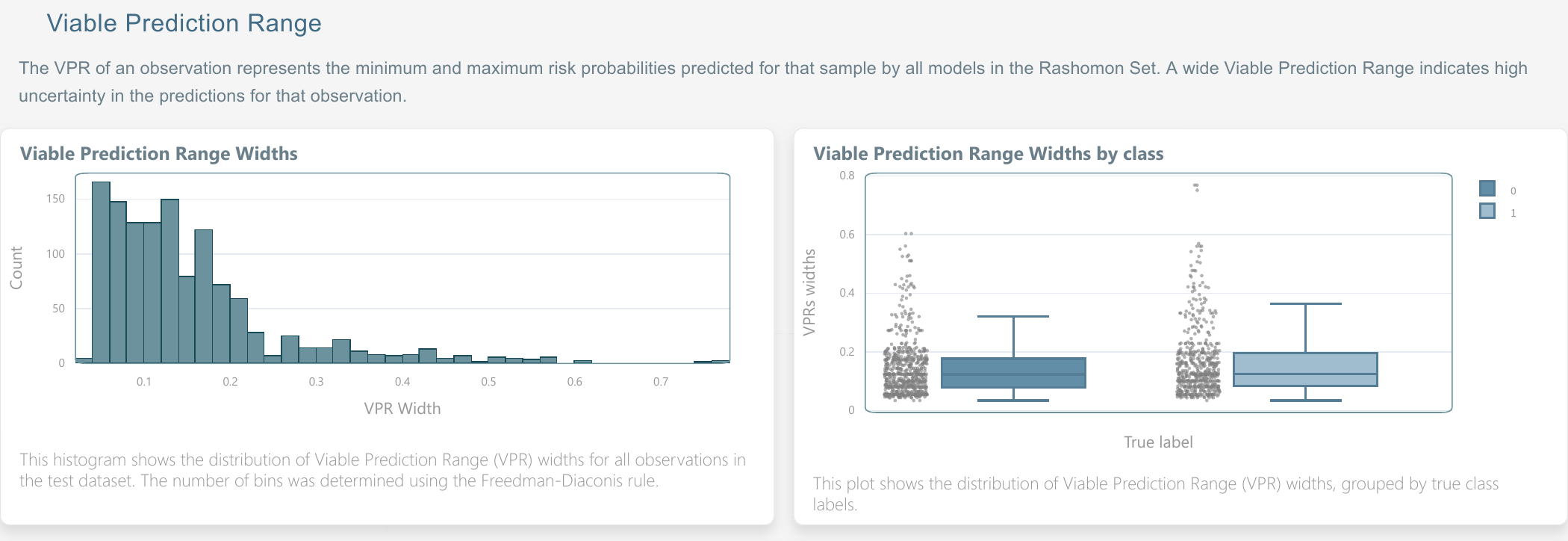} 
    \caption{Sample plots available on the Rashomon page of the application. The chart on the left is a histogram presenting the distribution of the VPR widths across all samples in the analyzed dataset, while the box plot on the right-hand side illustrates the widths with respect to the true labels.}
    \label{fig:rashomon2}
\end{figure}

After all parameters have been selected, an interactive dashboard with multiple charts is displayed, allowing a detailed analysis of the attributes and metrics related to the constructed Rashomon set. Additionally, users can switch between pages illustrating the Rashomon sets created using AutoGluon and H2O models with the same parameter configuration in order to explore the differences between these two frameworks. Figure \ref{fig:rashomon1} illustrates initial plots appearing on the dashboard after successfully constructing the Rashomon set. They contain primary characteristics of the constructed Rashomon set, such as the reference model, the base metric, and the number of included models. Additionally, Figure \ref{fig:rashomon2} presents sample plots available for analysis of the Viable Prediction Range metric calculated for each sample using the models from the Rashomon set. Similarly, all plots returned by the Visualizer class are included in the dashboard and available upon scrolling. 

\begin{figure}[!ht]
    \centering
    \begin{subfigure}[t]{0.8\textwidth}
        \includegraphics[width=\textwidth, keepaspectratio]{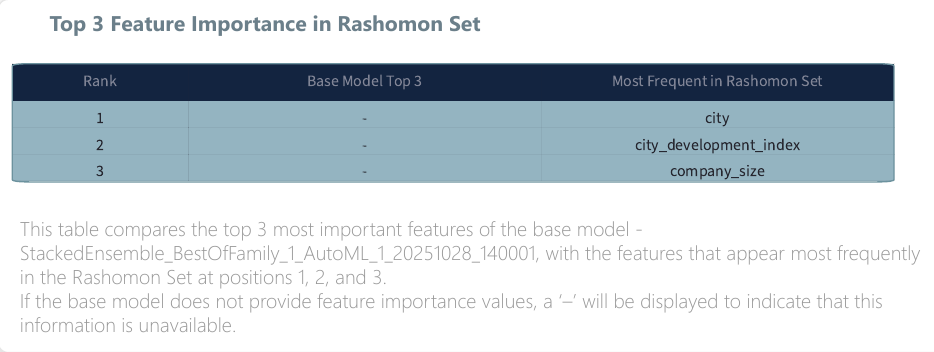}
        \caption{Table illustrating a scenario, when feature importance cannot be obtained from the reference model. In such cases, the first column of the table remains empty. }
        \label{table_fi}
    \end{subfigure}
    \par\bigskip
    \begin{subfigure}[t]{0.8\textwidth}
        \includegraphics[width=\textwidth, keepaspectratio]{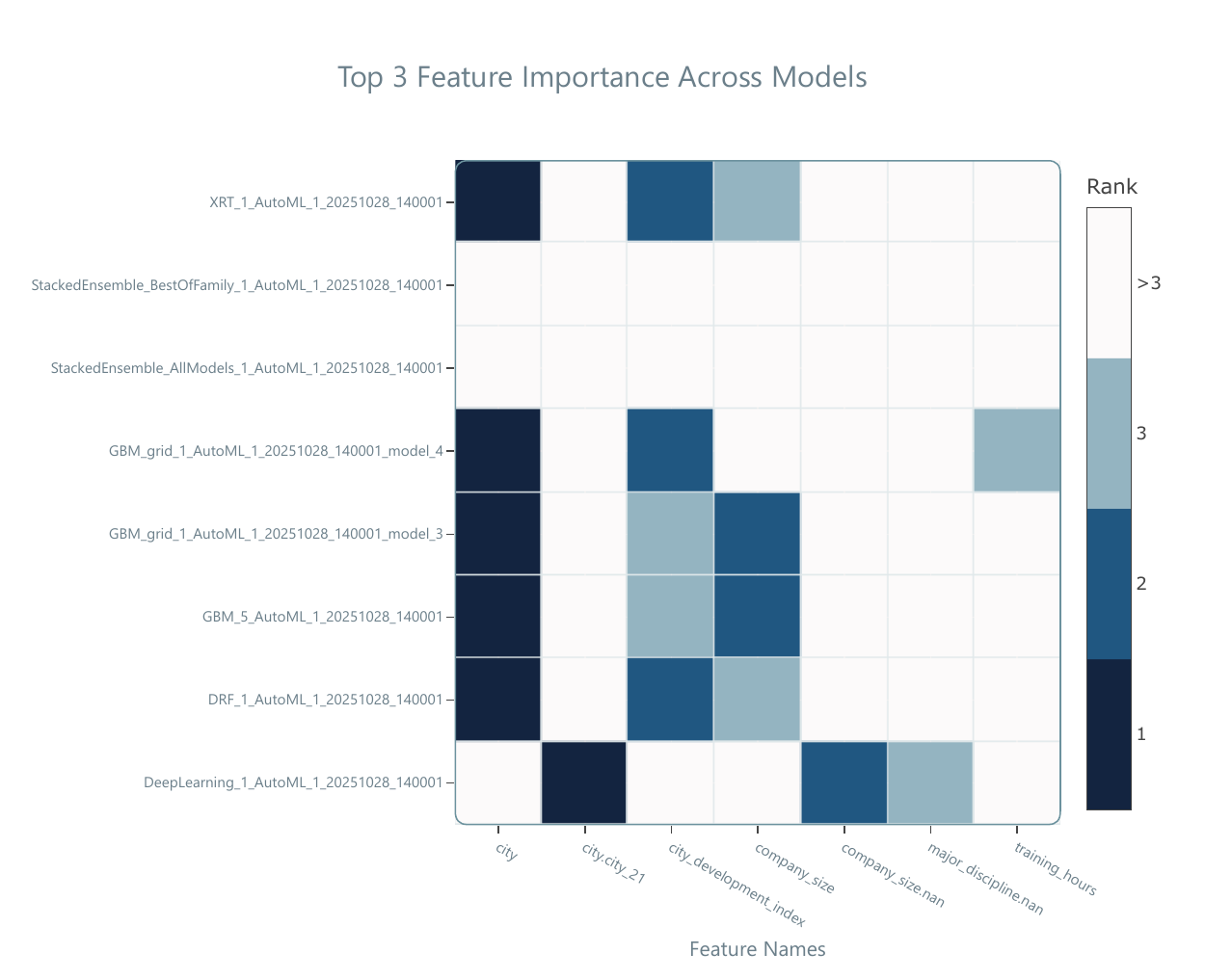}
        \caption{A heatmap presenting the top $3$ most important features obtained from every model in the Rashomon set. Regarding \texttt{StackedEnsembleBestOfFamily} and \texttt{StackedEnsembleAllModels} models, feature importances could not be obtained. Therefore, the corresponding matrix rows are empty.}
        \label{heatmap_fi}
    \end{subfigure}
    \label{all}
    \caption{Feature importance plots available on the dashboard.}
    \label{fig:rashomon3}
\end{figure}

\newpage
Figure \ref{fig:rashomon3} presents plots from the dashboard, related to feature importance obtained from the models in the Rashomon set. For some complex models, such as stacked ensembles, feature importance is not provided by the H2O framework. In such cases, a corresponding heatmap row remains empty, which is illustrated on the following charts. Figure \ref{table_fi} illustrates the top three most important features obtained from the reference model in comparison with the three features selected as the most important ones by the majority of the near-optimal models. Figure \ref{heatmap_fi} illustrates a heatmap enabling a more detailed analysis of these features for each model in the Rashomon set.

\begin{figure}[!ht]
    \centering
    \includegraphics[width=0.9\linewidth, keepaspectratio]{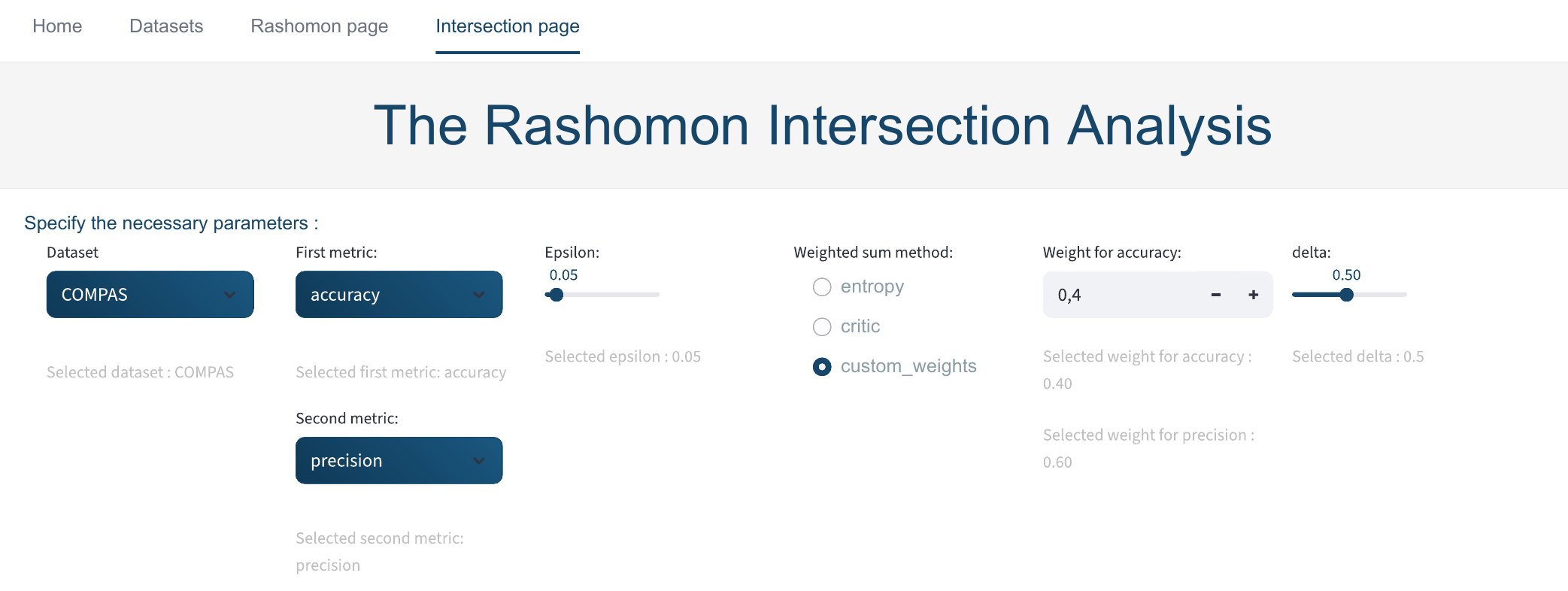} 
    \caption{The initial view of the Intersection page providing parameter selection using interactive widgets. Please note that after selecting the 'custom weights' weighted sum method, additional widgets for weights specification appear on the screen, allowing the selection of a weight for the first evaluation metric, while the second weight is calculated automatically as $1-w_1$.}
    \label{fig:intersection_widgets}
\end{figure}

\newpage
 \textbf{Intersection page}
    \\
    The last page of the application is dedicated to the analysis of the newly introduced concept of the Rashomon Intersection. Similar to the Rashomon page, users are asked to select a dataset and specify the necessary parameters, such as two evaluation metrics, epsilon and a weighted sum method ('custom weights', 'entropy', or 'CRITIC'). Figure \ref{fig:intersection_widgets} presents the initial view providing widgets for parameter selection. Similarly, after specifying certain parameters, the corresponding information is displayed below the widget.

\begin{figure}[!ht]
    \centering
    \includegraphics[width=0.9\linewidth, keepaspectratio]{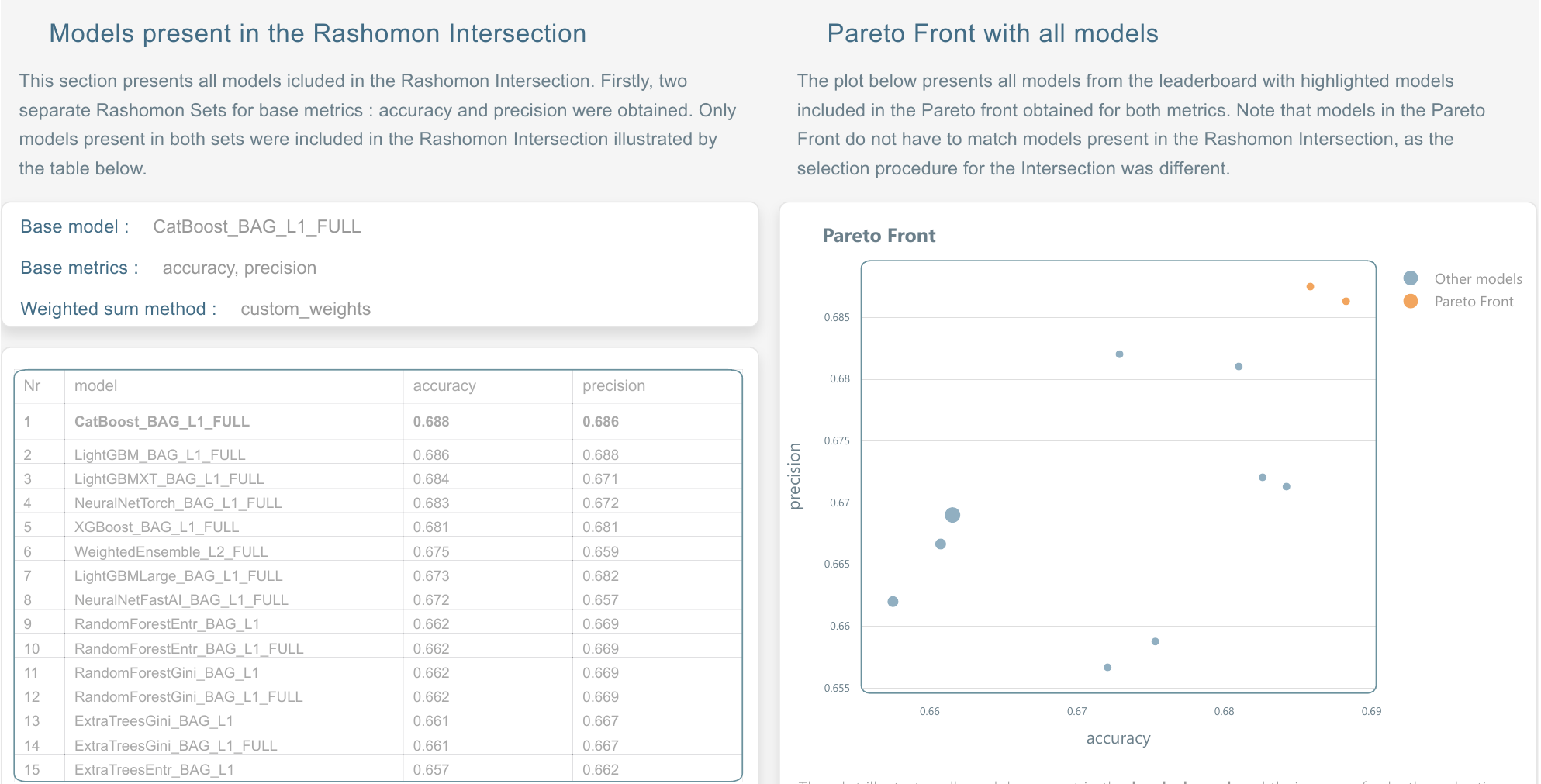} 
    \caption{The first part of the dashboard appearing on the Intersection page after the successful parameter specification and construction of the Rashomon Intersection. The table on the left-hand side presents models included in the Intersection (here for AutoGluon framework), along with their scores in both evaluation metrics. The right side of the dashboard focuses on the multi-objective optimization approach that computes the Pareto Front.}
    \label{fig:intersection1}
\end{figure}

\begin{figure}[!ht]
    \centering
    \includegraphics[width=\linewidth, keepaspectratio]{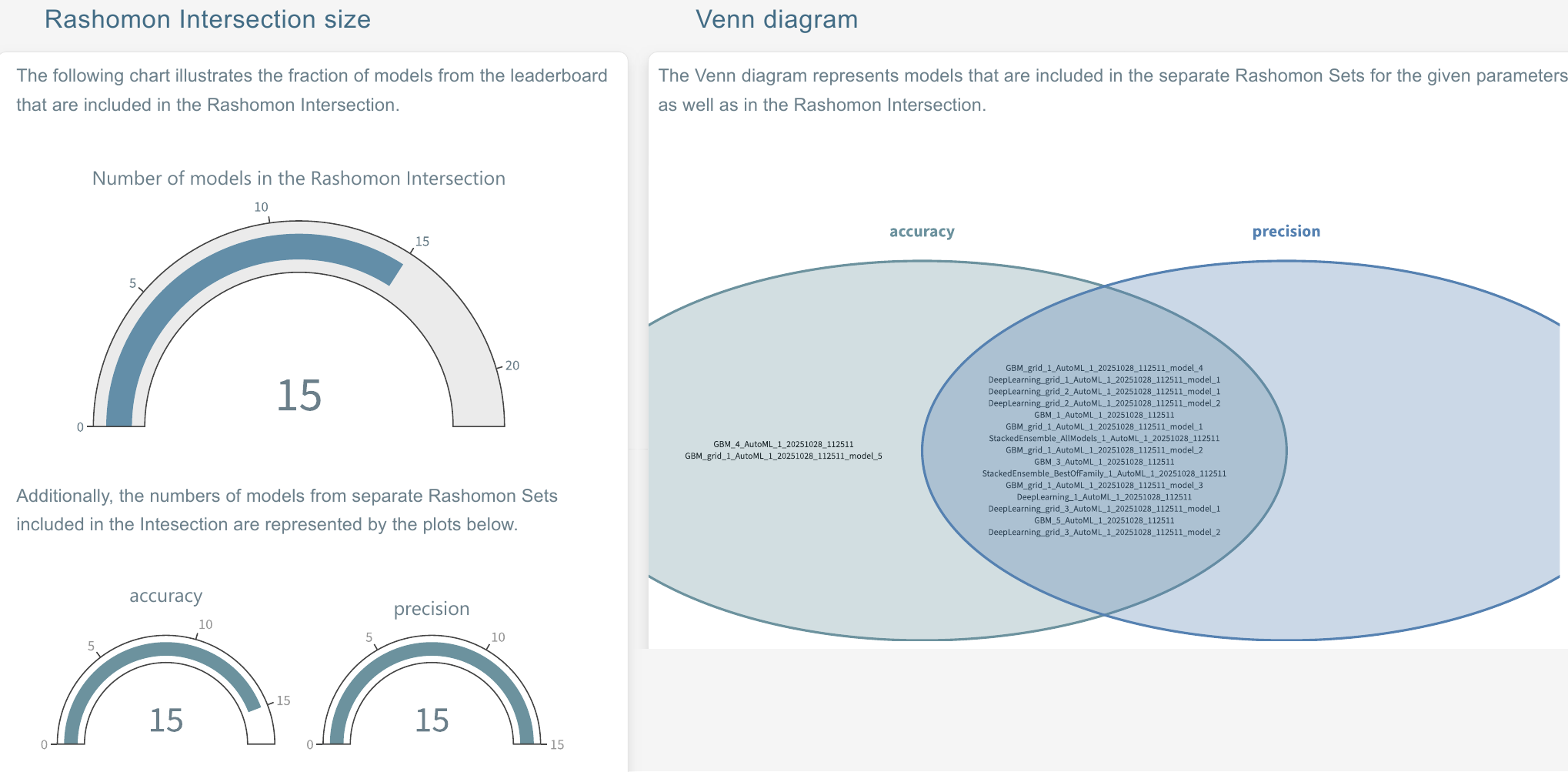} 
    \caption{Sample plots available on the Intersection page allowing the analysis of the separate Rashomon sets, as well as their intersection. Here we present an example scenario using the H2O framework. The Venn diagram and gauge plots illustrate which models did not achieve high enough precision score to be included in the Rashomon Intersection.}
    \label{fig:intersection2}
\end{figure}

    Then, a dashboard, resembling the one on the Rashomon page, is displayed. Since the metrics for the Rashomon set also apply to the Rashomon Intersection, many plots from the Rashomon page are reused on the Intersection page. Additionally, we enable the comparison between the properties of the Rashomon Intersection and a widely used concept of the Pareto Front. Figure \ref{fig:intersection1} presents the first part of the dashboard appearing on the Intersection page after the specification of all necessary parameters. Additionally, Figure \ref{fig:intersection2} presents sample plots available on the Intersection page that enable a brief analysis of the separate Rashomon sets (for both specified evaluation metrics) in comparison to their intersection. Please note that all plots visualizing metrics related to the Rashomon set, which are available on the Rashomon page, are also available for analysis of the Rashomon Intersection, as the predictive multiplicity metrics apply for both objects.
    
\section{Rashomon Intersection - selection of reference models}
\label{sec:weight_intersection}

To choose the weight values for selection of the reference model, we propose three methods: (1) the Custom weights method, (2) the Entropy method, and (3) the Criteria Importance Through Intercriteria Correlation (CRITIC) method.  Methods (2) and (3) are based on the distributions of the metric values $M_1$ and $M_2$ for each of the $m$  models in the Rashomon Intersection. Let $\hat{M_1} =[\hat{M_{11}},...\hat{M_{m1}}]$ denote a vector of $M_1$ values across all $m$ models. Similarly, let $\hat{M_2} =[\hat{M_{12}},...\hat{M_{m2}}]$ denote a vector of $M_2$ values across all $m$ models.

 \begin{enumerate}
        \item Custom weights - weight values are selected based on subjective preferences; when constructing the Rashomon Intersection, we can specify the importance of each metric for a given classification problem. For example, we can assign $w_1=0.3$ to metric $M_1$ and $w_2=0.7$ to metric $M_2$. 
        \item Entropy method - weights are calculated based on the theory of informational uncertainty (\cite{Wang2023}). Intuitively, a given objective (e.g., an evaluation metric $M$) will be assigned a higher weight if the differences in its values are more significant compared to those of other objectives (e.g, other metrics). For a better understanding of the Entropy method, let's consider the Rashomon Intersection consisting of $m$ models and two evaluation metrics $M_1$ and $M_2$ (i.e., objectives). The process of calculating weights $w_1$ and $w_2$ is as follows.
        
        \begin{enumerate}
            \item Normalize the values of $\hat{M_1}$ and $\hat{M_2}$ achieved by models using the sum normalization. For $j \in \{1,2\}$
            \begin{equation}
                    F_{ij} = \frac{\hat{M_{ij}}}{\sum_{k=1}^m\hat{M_{kj}}},
                \end{equation}  
            where $i\in [m]$ and $\hat{M_{ij}}$ is the value of the evaluation metric $M_j$ (here $M_1$ or $M_2$) achieved by the model $i$.
            
            \item For each metric, calculate its entropy as 
                \begin{equation}
                    E_j = -\frac{1}{ln(m)}\sum_{i=1}^m(F_{ij} \ ln(F_{ij})),
                \end{equation}  
                where $j\in\{1,2\}$, $m$ is the number of models in the Rashomon Intersection, and $F_{ij}$ is the normalized value of the evaluation metric $M_j$ for the model $i$. 
                \item Calculate weights for each metric as 
                \begin{equation}
                    w_j = \frac{1-E_j}{(1-E_1)+(1-E_2)}.
                \end{equation} 
                
        \end{enumerate}
        
        \item CRITIC method - by definition, weight values are computed based on the correlation matrix between objectives and the standard deviation of each objective (\cite{Wang2023}).

        To understand how weights are assigned for two criteria (here, two different evaluation metrics), we present the CRITIC formula for the case of two objectives.
        
        Assume we have the Rashomon Intersection consisting of $m$ models and two evaluation metrics $M_1$ and $M_2$ (i.e., objectives). The $\hat{M_1}$ and $\hat{M_2}$ are normalized the same as in the Entropy method.
        Let $\hat{\rho}$ denote the empirical correlation matrix between random variables $\hat{M_1}$ and $\hat{M_2}$. Formally, 
        \[
        \hat{\rho} =
        \begin{bmatrix}
        \hat{\rho}_{11} & \hat{\rho}_{12} \\
        \hat{\rho}_{21} & \hat{\rho}_{22} \\
        \end{bmatrix},
        \]
        where $\hat{\rho}_{11} = \hat{\rho}_{22} = 1$ (correlation of a random variable with itself) and 
        $\hat{\rho}_{12} = \hat{\rho}_{21}$ is the empirical correlation between $\hat{M}_1$ and $\hat{M}_2$.

        Additionally, let $\hat{\sigma}_1$, $\hat{\sigma}_2$ denote empirical standard deviations for $\hat{M_1}$ and $\hat{M_2}$, respectively. 
        
        Formally, to compute weights using the CRITIC method, we define:
        \begin{equation}
            c_1 = \hat{\sigma}_1 [(1-\hat{\rho}_{11}) +(1-\hat{\rho}_{12})],
        \end{equation}
        \begin{equation}
            c_2 = \hat{\sigma}_2 [(1-\hat{\rho}_{21}) +(1-\hat{\rho}_{22})].
        \end{equation}
        If $\hat{\sigma}_1=0$ or $ \hat{\sigma}_2=0$, the  correlation matrix $\hat{\rho}$ cannot be computed and the method fails.
        If $\hat{\sigma}_1, \hat{\sigma}_2$ > 0, then $\hat{\rho}_{11} = \hat{\rho}_{22} =1$ and the above formulas can be simplified to:
        \begin{equation}
            c_1 = \hat{\sigma}_1 (1-\hat{\rho}_{\hat{M1},\hat{M2}}), \quad
            c_2 = \hat{\sigma}_2 (1-\hat{\rho}_{\hat{M1},\hat{M2}}),
        \end{equation}
        where $\hat{\rho}_{M1,M2} = \hat{\rho}_{12} = \hat{\rho}_{21}$ denotes empirical correlation between $\hat{M_1}$ and $\hat{M_2}$.

        Weights using the CRITIC method are defined as:
        \begin{equation}
            w_1 = \frac{c_1}{c_1+c_2} = \frac{\hat{\sigma}_1 (1-\hat{\rho}_{\hat{M1},\hat{M2}})}{(\hat{\sigma}_1+\hat{\sigma}_2)(1-\hat{\rho}_{\hat{M1},\hat{M2}})} = \frac{\hat{\sigma}_1}{\hat{\sigma}_1+\hat{\sigma}_2},
        \end{equation}
         \begin{equation}
            w_2 = \frac{c_2}{c_1+c_2} = \frac{\hat{\sigma}_2 (1-\hat{\rho}_{\hat{M1},\hat{M2}})}{(\hat{\sigma}_1+\hat{\sigma}_2)(1-\hat{\rho}_{\hat{M1},\hat{M2}})} = \frac{\hat{\sigma}_2}{\hat{\sigma}_1+\hat{\sigma}_2},
        \end{equation}

        In this case, the weights depend only on the standard deviations of the metric values. 
        
        As mentioned above, this method is not always feasible. For instance, when either of the objectives has zero variance, the method fails. This may occur when all models present in the Rashomon Intersection produce the same values for one of the selected evaluation metrics. 
        
    \end{enumerate}

\section{Results across datasets}

\begin{table}[ht]
  \centering
  \caption{Reference model performance per dataset.}
  \label{tab:best_results}
  {\footnotesize
  \setlength{\tabcolsep}{4pt}
  \begin{tabular}{p{0.3cm} p{4.8cm} p{1.0cm} p{0.6cm} c c c c}
    \toprule
    \# & Dataset Name & OpenML ID & No.\ Feat.
      & \multicolumn{2}{c}{Accuracy}
      & \multicolumn{2}{c}{AUC} \\
    \cmidrule(lr){5-6} \cmidrule(lr){7-8}
    & & & & AG & H2O & AG & H2O \\
    \midrule
1 & APSFailure & 46908 & 171 & 0.9956 & 0.9945 & 0.9937 & 0.9917 \\
2 & Amazon\_employee\_access & 46905 & 10 & 0.9548 & 0.9498 & 0.8991 & 0.8635 \\
3 & Bank\_Customer\_Churn & 46911 & 11 & 0.8672 & 0.8604 & 0.8775 & 0.8751 \\
4 & Diabetes130US & 46922 & 48 & 0.9127 & 0.8862 & 0.6735 & 0.6709 \\
5 & E-CommereShippingData & 46924 & 11 & 0.6898 & 0.6658 & 0.7533 & 0.7549 \\
6 & Fitness\_Club & 46927 & 7 & 0.7973 & 0.7893 & 0.8377 & 0.8309 \\
7 & GiveMeSomeCredit & 46929 & 11 & 0.9389 & 0.9303 & 0.8703 & 0.8543 \\
8 & HR\_Analytics\_Job\_Change\_of \_Data\_Scientists & 46935 & 13 & 0.8006 & 0.7972 & 0.8103 & 0.8049 \\
9 & Is-this-a-good-customer & 46938 & 14 & 0.8886 & 0.8886 & 0.8015 & 0.8033 \\
10 & Marketing\_Campaign & 46940 & 26 & 0.9107 & 0.9018 & 0.9252 & 0.9211 \\
11 & NATICUSdroid & 46969 & 87 & 0.9557 & 0.9536 & 0.9882 & 0.9884 \\
12 & bank-marketing & 46910 & 14 & 0.8944 & 0.8777 & 0.7720 & 0.7761 \\
13 & blood-transfusion-service-center & 46913 & 5 & 0.8503 & 0.8503 & 0.8377 & 0.8330 \\
14 & churn & 46915 & 20 & 0.9672 & 0.9728 & 0.9406 & 0.9409 \\
15 & coil2000\_insurance\_policies & 46916 & 86 & 0.9409 & 0.9267 & 0.7813 & 0.7759 \\
16 & credit-g & 46918 & 21 & 0.7840 & 0.7800 & 0.8034 & 0.8140 \\
17 & credit\_card\_clients\_default & 46919 & 24 & 0.8240 & 0.8071 & 0.7957 & 0.7921 \\
18 & customer\_satisfaction\_in\_airline & 46920 & 22 & 0.9621 & 0.9606 & 0.9950 & 0.9948 \\
19 & diabetes & 46921 & 9 & 0.8281 & 0.8125 & 0.8836 & 0.8715 \\
20 & hazelnut-spread-contaminant-detection & 46930 & 31 & 0.9500 & 0.9517 & 0.9862 & 0.9877 \\
21 & heloc & 46932 & 24 & 0.7358 & 0.7285 & 0.8086 & 0.8104 \\
22 & in\_vehicle\_coupon\_recommendation & 46937 & 25 & 0.7846 & 0.7789 & 0.8505 & 0.8564 \\
23 & jm1 & 46979 & 22 & 0.8254 & 0.8012 & 0.7690 & 0.7691 \\
24 & online\_shoppers\_intention & 46947 & 18 & 0.9089 & 0.9056 & 0.9371 & 0.9372 \\
25 & polish\_companies\_bankruptcy & 46950 & 65 & 0.9689 & 0.9668 & 0.9674 & 0.9540 \\
26 & qsar-biodeg & 46952 & 42 & 0.9011 & 0.9053 & 0.9483 & 0.9579 \\
27 & seismic-bumps & 46956 & 16 & 0.9350 & 0.9350 & 0.8231 & 0.8024 \\
28 & taiwanese\_bankruptcy\_prediction & 46962 & 95 & 0.9765 & 0.9754 & 0.9662 & 0.9663 \\
    \bottomrule
  \end{tabular}}
\end{table}

\newpage

\section{Algorithm mapping}
\label{app:model_mapping}

\begin{table}[!ht]
  \centering
  \caption{Mapping of model name keywords to model type categories.}
  \label{tab:model_type_mapping}
  \small
  \begin{tabular}{llll}
    \toprule
    Keyword match & Example model names & Category & Framework \\
    \midrule
    \texttt{lightgbm}   & LightGBM, LightGBMLarge, LightGBMXT        & Gradient Boosting        & AG \\
    \texttt{xgboost}    & XGBoost, XGBoost\_BAG\_L1                  & Gradient Boosting        & AG \\
    \texttt{catboost}   & CatBoost, CatBoost\_BAG\_L2                & Gradient Boosting        & AG \\
    \texttt{gbm}        & GBM, GBM\_1                                & Gradient Boosting        & H2O \\
    \texttt{drf}        & DRF, DRF\_1                                & Tree Based Models        & H2O \\
    \texttt{extratrees} & ExtraTreesEntr, ExtraTreesGini             & Tree Based Models        & AG \\
    \texttt{forest}     & RandomForest, RandomForestEntr             & Tree Based Models        & AG \\
    \texttt{xrt}        & XRT, XRT\_1                                & Tree Based Models        & H2O \\
    \texttt{deeplearning} & DeepLearning, DeepLearning\_1            & Neural Networks          & H2O \\
    \texttt{neural}     & NeuralNetFastAI, NeuralNetTorch            & Neural Networks          & AG \\
    \texttt{glm}        & GLM, GLM\_1                                & Linear Models            & H2O \\
    \texttt{linear}     & LinearModel, LinearModel\_BAG\_L1         & Linear Models            & AG \\
    \texttt{ensemble}   & WeightedEnsemble\_L3, StackedEnsemble\_AllModels & Model Ensemble     & AG, H2O \\
    (no match)          & KNeighbors, NaiveBayes                     & Other                    & AG, H2O \\
    \bottomrule
  \end{tabular}
\end{table}

\newpage

\section{Limitations}
\label{sec:limitations}

Several limitations of the present study should be acknowledged.

\textbf{Scope restricted to binary classification.}
All experiments are conducted exclusively on binary classification tasks retrieved from OpenML. Whether the observed differences in Rashomon set structure between AutoGluon and H2O generalise to multiclass classification, regression, or other prediction tasks remains an open question. The metrics employed---ambiguity, discrepancy, VPR, and Rashomon Capacity---admit natural multiclass extensions~\citep{capacity}, but their empirical behaviour in those settings has not been examined here.

\textbf{Framework configuration asymmetry.}
AutoGluon was configured with the \textit{good\_quality} preset, which employs bagging and stacking and can produce a large pool of reference models, while H2O AutoML was capped at 20 models. Although both frameworks were given an equivalent time budget of 2h per fold, the ceiling on H2O's model count is a structural confound: observed differences in Rashomon Ratio and set size may partly reflect the number of models available rather than the intrinsic diversity of each framework's hypothesis space. Future work should control for model count or report results stratified by pool size.

\textbf{Inconsistency in feature importance estimation.}
For AutoGluon, feature importance is obtained directly from the predictor, whereas for H2O stacked ensembles---which do not natively expose compatible importance scores---permutation importance is computed post-hoc using \texttt{scikit-learn}, scored by accuracy. Non-ensemble H2O models use framework-internal estimates. Because the XHack analysis and RQ3 conclusions rest on feature importance rankings, this methodological inconsistency introduces a potential confound: measured differences in explanation hackability between frameworks may partially reflect differences in how importance is computed rather than genuine differences in model behaviour.

\textbf{Rashomon sets bounded by AutoML output.}
The Rashomon sets analyzed here are restricted to models that the AutoML framework chose to train and retain. Both AutoGluon and H2O return only a subset of all models explored during the search process, biased towards high-performing configurations. This means the empirical Rashomon set is a subset of the theoretical one, and conclusions about set size and diversity are conditional on each framework's internal model selection and pruning strategy.

\textbf{Single time budget.}
All experiments were conducted under a fixed time limit of 2h per fold for both frameworks. This choice reflects a single operating point in the trade-off between computational cost and model pool diversity. It is plausible that the observed differences in Rashomon set size and structure between AutoGluon and H2O are sensitive to this budget.

\textbf{ARSA ML framework coverage.}
The current implementation of ARSA ML provides native support for AutoGluon and H2O only. Although a custom converter interface exists for other frameworks, the effort required for integration may limit adoption. Results and tooling cannot be directly applied to other widely-used AutoML systems, such as auto-sklearn or FLAML, without additional engineering.

\section{Negative societal impact}
\label{app:societal_impact}
 By quantifying and publicly reporting x-hackability scores across frameworks and datasets, this work  provides information of where explanation manipulation is structurally easiest. Our findings demonstrate that H2O's Rashomon sets consistently exhibit higher and more variable x-hackability than those of AutoGluon. While this result is intended to alert practitioners and auditors to framework-level risk, it could equally guide a malicious actor in selecting a framework and tolerance threshold that maximizes the availability of models supporting a desired narrative.
 
\section{Compute Resources}
\label{app:resources}
Experiments described in Section~\ref{sec:results} were computed on a cluster consisting of 2× AMD EPYC 7413 CPUs (48 cores), and 3TB of RAM. Each pipeline execution was constrained to a maximum memory usage of 32 GB and 8 cores.

\end{document}